%% file: arxiv.tex
\documentclass[10pt,twocolumn,letterpaper]{article}

\usepackage[table]{xcolor}
\usepackage{color}
\definecolor{mygreen}{RGB}{146, 199, 113} 
\definecolor{mypurple}{RGB}{146, 199, 113}

\usepackage{cvpr}              
\usepackage{url}
\usepackage{algorithm}
\usepackage{algorithmic}

\usepackage{multirow}

\usepackage{graphicx}
\usepackage{adjustbox}
\usepackage{amsmath}
\usepackage{array}
\usepackage{booktabs}
\usepackage{amssymb}
\usepackage{dsfont}
\usepackage{graphicx}
\usepackage{footmisc}
\usepackage{makecell}
\usepackage[accsupp]{axessibility}


\definecolor{cvprblue}{rgb}{0.21,0.49,0.74}
\usepackage[pagebackref,breaklinks,colorlinks,allcolors=cvprblue]{hyperref}

\def\paperID{3076} 
\def\confName{CVPR}
\def\confYear{2025}

\title{
    \centering
    WAVE: Weight Templates for Adaptive Initialization of Variable-sized Models
}

\author{Fu Feng\textsuperscript{\rm 1,2}\thanks{Equal contribution}\quad Yucheng Xie\textsuperscript{\rm 1,2}\footnotemark[1]\quad Jing Wang\textsuperscript{\rm 1,2}\thanks{Corresponding authors}\quad Xin Geng\textsuperscript{\rm 1,2}\footnotemark[2]\\
\textsuperscript{\rm 1}School of Computer Science and Engineering, Southeast University, Nanjing, China\\
\textsuperscript{\rm 2}Key Laboratory of New Generation Artificial Intelligence Technology and Its Interdisciplinary \\Applications (Southeast University), Ministry of Education, China\\
{\tt\small \{fufeng, xieyc, wangjing91, xgeng\}@seu.edu.cn}\\
}

\begin{document}
\maketitle

\begin{abstract}
The growing complexity of model parameters underscores the significance of pre-trained models. 
However, deployment constraints often necessitate models of varying sizes, exposing limitations in the conventional pre-training and fine-tuning paradigm, particularly when target model sizes are incompatible with pre-trained ones.
To address this challenge, we propose WAVE, a novel approach that reformulates variable-sized model initialization from a multi-task perspective, where initializing each model size is treated as a distinct task. 
WAVE employs shared, size-agnostic weight templates alongside size-specific weight scalers to achieve consistent initialization across various model sizes.
These weight templates, constructed within the \textit{Learngene} framework, integrate knowledge from pre-trained models through a distillation process constrained by Kronecker-based rules.
Target models are then initialized by concatenating and weighting these templates, with adaptive connection rules established by lightweight weight scalers, whose parameters are learned from minimal training data.
Extensive experiments demonstrate the efficiency of WAVE, achieving state-of-the-art performance in initializing models of various depth and width. The knowledge encapsulated in weight templates is also task-agnostic, allowing for seamless transfer across diverse downstream datasets. Code will be made available at~\href{https://github.com/fu-feng/WAVE}{https://github.com/fu-feng/WAVE}.
\end{abstract} 

\section{Introduction}
With the exponential increase in model parameters, the demands for time, energy, and computational resources during training have escalated significantly~\cite{touvron2023llama, achiam2023gpt}. Training a model from scratch for new tasks, especially those with Transformer architectures like Vision Transformers (ViTs)~\cite{dosovitskiy2020image}, is inefficient. As a result, fine-tuning pre-trained models has become the preferred approach~\cite{liu2021swin, wu2021cvt, li2024svasp}.
However, in practical applications, model deployment is often constrained by various factors, such as memory usage, processing power, and response time~\cite{zhang2022minivit}, necessitating models of \textit{variable sizes}. 
In contrast, most off-the-shelf pre-trained models, such as the 12-layer ViT-B models~\cite{touvron2021training}, are available in only a limited set of sizes. 
As a result, target models that do not align with these predefined sizes require re-pretraining on large datasets before deployment.

\begin{figure}[t]
  \centering
  \includegraphics[width=\linewidth]{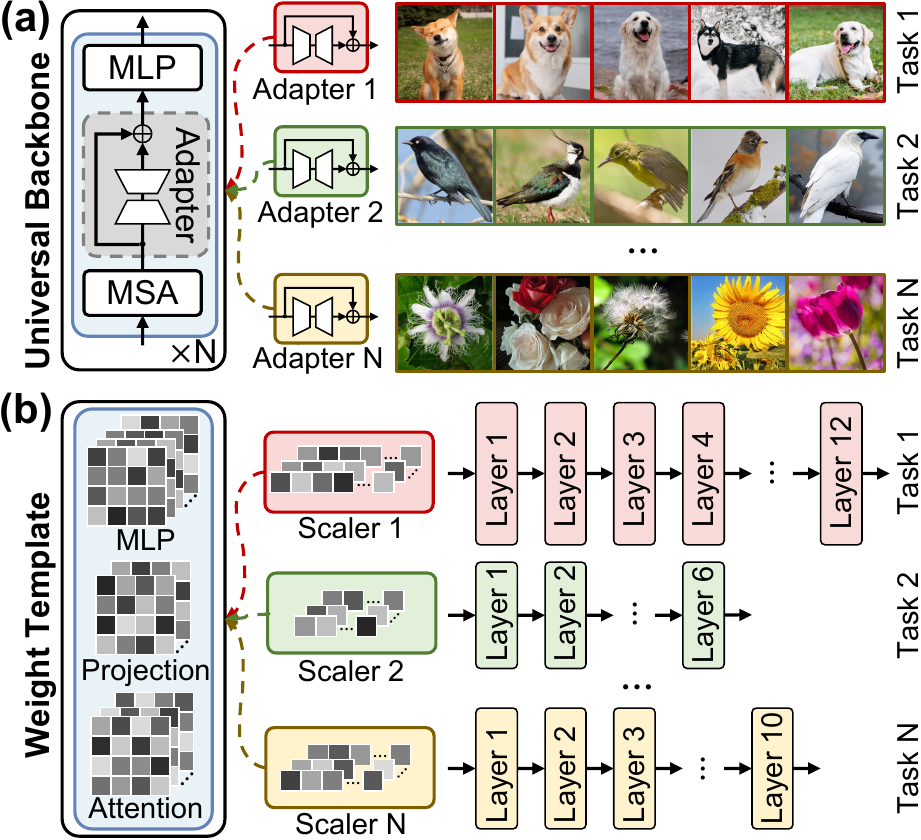}
  \vspace{-0.15in}
  \caption{(a) Multi-task learning typically relies on a universal backbone with \textbf{\textit{task-agnostic}} knowledge, complemented by a few trainable adapters for \textbf{\textit{task-specific}} adaptation. 
  (b) WAVE reformulates variable-sized model initialization as a multi-task problem by treating the initialization of each model size as a distinct task. By doing so, WAVE employs shared weight templates encapsulating \textbf{\textit{size-agnostic}} knowledge, along with a few trainable weight scalers for \textbf{\textit{size-specific}} initialization across various model sizes.}
  \label{fig:motivation}
\end{figure}

Several methods~\cite{wang2023learning, xu2023initializing} have explored leveraging pre-trained models to initialize target models with mismatched sizes by selecting or transforming pre-trained weight matrices. 
However, these methods often disrupt the structured knowledge embedded in original models or introduce excessive random parameters, limiting their effectiveness.
By abstracting the initialization of models with specific sizes as independent tasks, the challenge of initializing variable-sized models can be reformulated as a multi-task problem.
Multi-task learning typically requires only a minimal number of learnable parameters for rapid adaptation to new tasks (e.g., Lora~\cite{hu2021lora}), enabled by a universal backbone, encapsulating task-agnostic knowledge~\cite{triantafillou2021learning}, as illustrated in Figure~\ref{fig:motivation}.
Motivated by this, we can ask: \textit{Can a set of universal weight templates enable the initialization of variable-sized models with only a small number of learnable parameters?}

Recently, \textit{Learngene}~\cite{wang2023learngene, feng2023genes} has emerged as a novel framework for model initialization based on pre-trained models, referred to as ancestry models. Unlike existing methods that directly utilize pre-trained models for initialization~\cite{xu2023initializing, wang2023learning}, \textit{Learngene} introduces an additional step by integrating and encapsulating pre-trained knowledge into compact neural fragments, termed learngenes. This \textit{\textbf{once-for-all}} process extracts size-agnostic knowledge that enables the initialization of target models with variable sizes.

Building on the \textit{Learngene} framework, we propose WAVE, a novel and flexible approach for model initialization. 
WAVE employs shared \textit{weight templates} that encapsulate size-agnostic knowledge as learngenes.
These templates structurally integrate knowledge from pre-trained models through a distillation process facilitated by an auxiliary model~\cite{xia2024transformer}.
Downstream models initialize their weight matrices by concatenating and weighting these templates using the Kronecker product, with adaptive connection rules determined by lightweight weight scalers containing only a few thousand parameters. 
During initializing, only weight scalers need to be tailored to the target model sizes, and their parameters can be efficiently trained with limited data.

Integrating pre-trained knowledge into weight templates is a \textit{\textbf{once-for-all}} process requiring 150 epochs on ImageNet-1K, after which model initialization involves training only lightweight scalers while keeping the templates fixed.
Extensive experiments validate WAVE’s state-of-the-art performance in initializing variable-sized models across diverse datasets. 
Notably, with only 3.3\% of pre-trained model (i.e., ancestry model) parameters transferred, models initialized by WAVE and trained for just 10 epochs outperform those directly pre-trained for 150 epochs, achieving $15n\times$ computational savings for $n$ variable-sized models. 
Visualizations further reveal that the size-agnostic knowledge in weight templates is highly structured, consistent with patterns observed in pre-trained models~\cite{xia2024transformer, trockman2023mimetic}.

Our main contributions are as follows: 
1) We propose WAVE, a novel method that approaches model initialization from a multi-task perspective. 
WAVE enhances flexibility and efficiency by constructing shared weight templates, termed learngenes, combined with lightweight weight scalers trained to initialize target models.
2) We introduce the first comprehensive benchmark for learngenes, systematically evaluating their initialization and transfer capabilities.
3) Extensive experiments validate the effectiveness of WAVE, which achieves state-of-the-art performance compared to other model initialization and learngene methods. Moreover, we demonstrate that the knowledge encapsulated in weight templates is highly structured, enabling efficient and robust initialization.

\section{Related Work}
\subsection{Model Initialization}
Model initialization is crucial for the convergence speed and final performance of neural networks~\cite{arpit2019initialize, huang2020improving}. Traditional methods typically apply pre-set rules to random parameters~\cite{glorot2010understanding, chen2021empirical}. 
The advent of pre-trained models has advanced initialization methods, with fine-tuning emerging as the standard practice for many applications~\cite{zoph2020rethinking}. 
However, the fixed sizes and architectures of pre-trained models often limit their flexibility, highlighting the need for more versatile initialization methods.

Several methods have explored direct model initialization through observed parameter patterns or hypernetworks. 
Mimetic Initialization~\cite{trockman2023mimetic} leverages parameter patterns identified from attention heads in pre-trained models to initialize new ones, while GHN3~\cite{knyazev2023canwescale} predicts target model parameters using a graph hypernetwork.
Other methods transfer parameters from pre-trained models to target models, such as Weight Selection~\cite{xu2023initializing}, which transfers selected weights from larger models to smaller ones, and LiGO~\cite{wang2023learning}, which scales smaller models to initialize larger ones.

Despite these advancements, these methods often incur high transfer costs and risk inefficiency or negative transfer due to parameter mismatches. 
Our WAVE overcomes these limitations by constructing shared weight templates, which integrate pre-trained knowledge to improve transferability and scalability across models of varying sizes.

\subsection{Learngene}
\textit{Learngene} represents an innovative approach to model initialization and knowledge transfer, inspired by the biological process where our brains are initialized by evolved genes rather than through random initialization~\cite{wang2023learngene, feng2023genes, zador2019critique}. Learngenes are compact neural network fragments that encapsulate only size-agnostic knowledge, making them lightweight and easily transferable, much like genes in nature~\cite{feng2024transferring}. 
This distinguishes learngenes from mainstream methods, such as direct fine-tuning of pre-trained models~\cite{zoph2020rethinking, chakraborty2022efficient} or transforming pre-trained models to adapt to variable-sized models~\cite{wang2023learning, xu2023initializing, xia2024initializing}.

Current learngene methods primarily focus on a layer-by-layer basis. For instance, Heur-LG~\cite{wang2022learngene} selects layers with minimal gradient changes as learngenes during continuous learning, while Auto-LG~\cite{wang2023learngene} considers that only layers with representations similar to target networks should be transferred. TLEG~\cite{xia2024transformer} characterizes model initialization as a linear combination of two layers. 
Additionally, KIND~\cite{xie2024kind} and FINE~\cite{xie2024fine} apply singular value decomposition (SVD) to each weight matrix, enabling the sharing of singular vectors for initializing models of varying sizes.
However, existing learngenes only address depth variations.

In contrast, WAVE introduces a novel form of learngenes, overcoming layer-based limitations by treating each weight matrix as a concatenation and weighted combination of weight templates.
This method enhances the flexibility of learngenes by incorporating weight scalers, thus enabling scalable initialization for variable-sized models.

\section{Methods}
\begin{figure*}[tb]
  \centering
  \includegraphics[width=\linewidth]{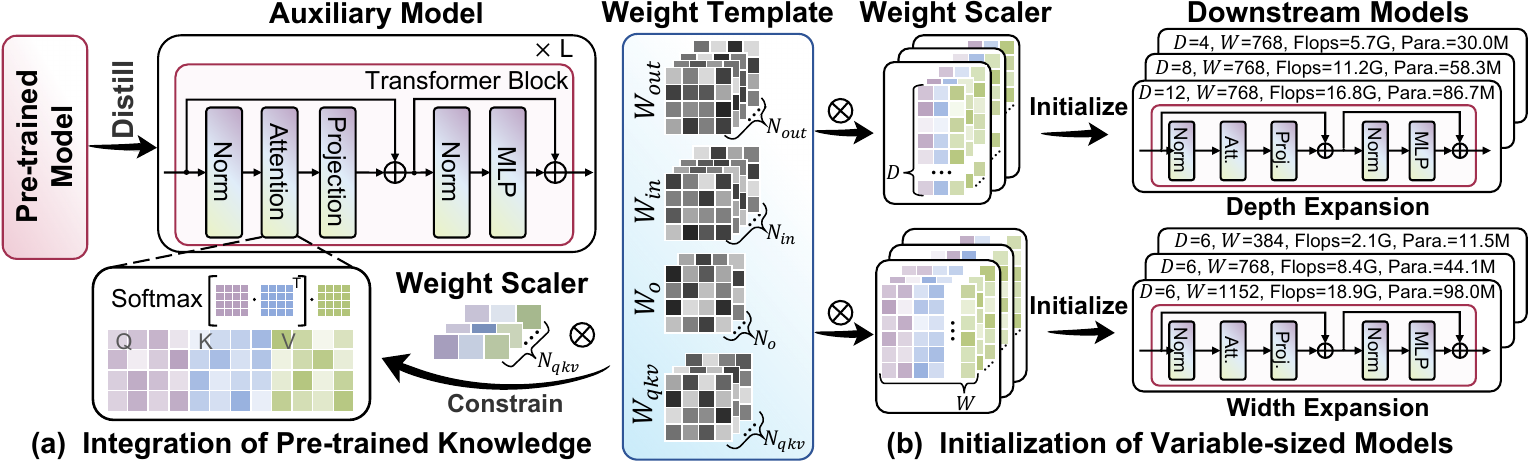}
  \caption{(a) The knowledge in pre-trained models is integrated into structured knowledge within weight templates through distillation, assisted by an auxiliary model constrained by the rules in Eq.\eqref{equ:kro}, where $\otimes$ is the Kronecker product. 
  (b) For initializing models of variable sizes, only the corresponding weight scalers are initialized according to the target model size. These scalers are trained with a small amount of data to learn connection rules of templates, while the weight templates remain frozen to retain the structured knowledge.}
  \label{fig:main}
\end{figure*}

WAVE is an innovative \textit{Learngene} method that constructs shared weight templates, termed learngenes, to initialize models of variable sizes.
This section begins by introducing how weight templates and scalers reconstruct neural network parameters using Kronecker products.
It then details the integration of pre-trained knowledge into structured, size-agnostic knowledge within weight templates, followed by the procedure for initializing models of variable sizes using these weight templates.
\subsection{Preliminaries}
The Vision Transformer (ViT) comprises an encoder stacking $L$ layers, each containing a multi-head self-attention (MSA) mechanism and a multi-layer perception (MLP). In each layer, a single-head self-attention $A_i$ is composed of a query $Q_i$, key $K_i$ and value $V_i\in \mathbb{R}^{N\times d}$ with parameter matrices $W_q^i$,  $W_k^i$ and $W_v^i\in \mathbb{R}^{D\times d}$, performing self-attention:
\begin{equation}
    A_i = \text{softmax}(\frac{Q_i K_i^\top}{\sqrt{d}})V_i\,,\; A_i\in \mathbb{R}^{N\times d}
\end{equation}
where $N$ is the number of patches, $D$ is the dimensional of patch embeddings, and $d$ is the projected dimension for each attention head. 

MSA processes information from $h$ attention heads and uses a weight matrix $W_{o}$ to project the concatenated outputs:
\begin{equation}
    \text{MSA} = \text{concat}(A_1, A_2, ..., A_{h}) W_{o}\,,\;W_{o}\in \mathbb{R}^{hd\times D}
\label{equ:msa}
\end{equation}
In the implementation of MSA, the parameter matrices $W_q$, $W_k$ and $W_v\in \mathbb{R}^{D\times d}$ for $h$ attention heads are typically combined into a single parameter matrix $W_{qkv}\in \mathbb{R}^{D\times 3hd}$.

The MLP consists of two linear transformations, $W_{in}\in \mathbb{R}^{D\times D'}$ and $W_{out}\in \mathbb{R}^{D'\times D}$, with a GELU~\cite{hendrycks2016gaussian} activation, computed as:
\begin{equation}
    \text{MLP}(x)=\text{GELU}(xW_{in}+b_1)W_{out} + b_2
\label{equ:mlp}
\end{equation}
where $b_1$ and $b_2$ are the bias for the linear transformations, and $D'$ is the dimension of the hidden layers.

\subsection{Weight Template}
\label{sec:wt}
\textbf{Foundations for the Existence of Weight Templates.} Evidence suggests the existence of shared templates among diverse weight matrices. For instance, Template Kernels~\cite{liu2022polyhistor} demonstrate the ability to generate diverse adapters for various tasks, while the Universal Template~\cite{triantafillou2021learning} enhances generalization in few-shot learning scenarios.
Additionally, studies on pre-trained Vision Transformers (ViTs) provide additional insights. Mimetic Initialization~\cite{trockman2023mimetic} reveals significant correlations among the weights of self-attention layers, and TLEG~\cite{xia2024transformer} identifies linear relationships among the weights of different layers. 

Building on these insights, WAVE utilizes shared weight templates encapsulating size-agnostic knowledge as learngenes. 
The initialization of variable-sized models is flexibly achieved by applying size-specific weight scalers to these templates via Kronecker products, as illustrated in Figure~\ref{fig:main}.

The main weight matrices in ViTs with $L$ layers are represented as $\theta = \{W_{qkv}^{(1\thicksim L)}, W_{o}^{(1\thicksim L)}, W_{in}^{(1\thicksim L)}, W_{out}^{(1\thicksim L)}\}$\footnote{$W_{qkv}^{(1\thicksim L)}$ denotes the set $\{W_{qkv}^{(1)}, W_{qkv}^{(2)}, \dots, W_{qkv}^{(L)}\}$. Similar notations throughout the paper follow this convention.}. 
These matrices can be reconstructed using a unified set of shared weight templates $\mathcal{T} = \{T_{qkv}^{(1\thicksim N_{qkv})}, T_{o}^{(1\thicksim N_o)}, T_{in}^{(1\thicksim N_{in})}, T_{out}^{(1\thicksim N_{out}})\}$\footnote{Parameters in normalization, bias and image relative position encoding (irpe) account for a small part and have little impact when randomly initialized, as detailed in Table~\ref{tab:ablation}.}. 
To effectively leverage the structured information in weight templates, WAVE initializes the corresponding weight matrices by concatenating and weighting these templates using the Kronecker product, facilitated by Layer-Wise Scaling Kernels~\cite{liu2022polyhistor}. 
Let $W_{\star}^{(l)}$ represent any weight matrix in layer $l$, and $T_{\star}^{(t)}$ denote the $t$-th corresponding weight template, where $\star \in \{qkv, o, in, out\}$. The $W_{\star}^{(l)}$ is initialized as follows:
\begin{equation}
    W_{\star}^{(l)} = \sum_{t=1}^{N_{\star}} T_{\star}^{(t)} \otimes S^{(l, t)}_{\star}
\label{equ:kro}
\end{equation}
\begin{equation*}
    W_{\star}^{(l)}\in \mathbb{R}^{m_1\times m_2}\,,\; T_{\star}^{(t)}\in \mathbb{R}^{w_1\times w_2}\,,\; S^{(l, t)}_{\star}\in \mathbb{R}^{\frac{m_1}{w_1}\times \frac{m_2}{w_2}}
\end{equation*}
where $\otimes$ denotes the Kronecker product. $m_1$ and $m_2$ represent the dimensions of weight matrices, while $w_1$ and $w_2$ determine the size of weight templates. 
$S_{\star}^{(l, t)}$ is the weight scaler performing the Kronecker product with  $T_{\star}^{(t)}$. The set of all trainable weight scalers is denoted as $\mathcal{S}=\{S_{qkv}^{(1\thicksim L, 1\thicksim N_{qkv})}, S_{o}^{(1\thicksim L, 1\thicksim N_{o})}, S_{in}^{(1\thicksim L, 1\thicksim N_{in})}, S_{out}^{(1\thicksim L, 1\thicksim N_{out})}\}$. 
To succinctly represent the Kronecker product rule in Eq.~\eqref{equ:kro}, we abbreviate it as:
\begin{equation}
    \theta = \mathcal{T}\otimes \mathcal{S}
\label{equ:ob_sip}
\end{equation}
We also demonstrate that layer-based learngenes, including Heur-LG~\cite{wang2022learngene}, Auto-LG~\cite{wang2022learngene}, and TLEG~\cite{xia2024transformer}, are special cases of WAVE. A detailed proof is provided in Appendix~\ref{sec:proof}.

\subsection{Integration of Pre-trained Knowledge}
\label{sec:know_condense}
The weight matrices of models can be initialized using weight templates and scalers via Kronecker product, enabling robust initialization by effectively leveraging the structured knowledge encapsulated within the weight templates.
Next, we outline the process of integrating knowledge from pre-trained models into these structured weight templates, following the rules defined in Eq.\eqref{equ:kro}.

To transfer knowledge from pre-trained models into weight templates, we employ knowledge distillation~\cite{hinton2015distilling}, a widely-used method in knowledge transfer. 
Since weight templates themselves lack the capacity to learn from data independently, we construct an auxiliary model $f_{\text{aux}}(\theta_{\text{aux}})$~\cite{xia2024transformer}, whose parameters are derived from weight templates and scalers under the rule defined in Eq.\eqref{equ:kro}. 
The parameter updates of the auxiliary model are achieved by updating weight templates and scalers (see Algorithm~\ref{alg:algorithm} for details), following the optimization objective:
\begin{equation}
    {\underset{\mathcal{T}, \mathcal{S}_{\text{aux}}}{{\arg\min}} \, \mathcal{L}(f_{\text{aux}}(\theta_{\text{aux}}, x), y)}, \quad
    \text{s.t.}\, \theta_{\text{aux}} = \mathcal{T}\otimes \mathcal{S}_{\text{aux}}
\label{equ:ob}
\end{equation}
where $\mathcal{T}$ is the target set of shared weight templates for universally initializing variable-sized models. $\mathcal{S}_{\text{aux}}$ represents the set of weight scalers specific to the auxiliary model $f_{\text{aux}}$, which are discarded after the knowledge integration.

In this manner, the auxiliary model serves as both a medium for transferring knowledge from pre-trained models to weight templates, and a bottleneck for filtering out unstructured knowledge lacking size-agnostic properties required for initializing variable-sized downstream models.

The learning process for weight templates adopts soft distillation loss and classification loss, computed as:
\begin{equation}
    \mathcal{L} = \text{KL}(z_{\text{pre}}||z_{\text{aux}}) + \text{CE}(z_{\text{aux}}, y)
\label{equ:loss}
\end{equation}
where $\text{KL}(\cdot, \cdot)$ denotes the KL-divergence loss, with $z_{\text{pre}}$ and $z_{\text{aux}}$ being the logits outputs of the pre-trained model and auxiliary model. 
$\text{CE}(\cdot, \cdot)$ denotes the cross-entropy loss, and $y$ is the ground-truth label of the image.
Note that $\mathcal{L}$ is exclusively used to update the parameters of weight templates and weight scalers (Eq.~\eqref{equ:ob}), while the parameters of the auxiliary model are indirectly updated by being reconstructed under the rule in Eq.~\eqref{equ:kro} at each iteration.

\subsection{Initialization of Variable-sized Models}
\label{sec:know_decom}
We have successfully constructed weight templates by integrating knowledge from the pre-trained model. 
Next, we address the process of initializing variable-sized models using these weight templates $\mathcal{T}$. 
Existing learngene methods typically rely on predetermined strategies, such as linear expansion~\cite{xia2024transformer} or manual selection~\cite{xia2024initializing}, which considerably limit their adaptability to models of variable sizes.

Drawing inspiration from trainable adapters in addressing multi-tasking challenges, we \textbf{\textit{freeze}} the weights of weight templates $\mathcal{T}$ and then construct the weight scalers $\mathcal{S}$ tailored to the target model size, thereby enabling adaptation to networks of varying sizes. Further details on the initialization of weight scalers are provided in Appendix~\ref{app:init_ws}.

Specifically, to initialize a target model $f_{\text{tar}}$ with parameters $\theta_{\text{tar}}$, each weight matrix $W_{\star, \text{tar}}^{(l)}\in \mathbb{R}^{m_1\times m_2}$ in $\theta_{\text{tar}}$ requires an associated set of weight scalers.
Their dimensions, determined by the size of the weight matrix being initialized and the associated weight templates, are given by $S_{\star,\text{tar}}^{(l,1\thicksim N_{\star})}\in \mathbb{R}^{N_{\star} \times \frac{m_1}{w_1}\times \frac{m_2}{w_2}}$ according to Eq.~\eqref{equ:kro}, where $w_1$ and $w_2$ are the dimensions of the weight templates $T_{\star}^{(1\thicksim N_{\star})} \in \mathbb{R}^{N_{\star} \times w_1\times w_2}$.

A small amount of data is then used to train the weight scalers $\mathcal{S_{\text{tar}}}$ with the following objective: 
\begin{equation}
    {\underset{\mathcal{S}_{\text{tar}}}{{\arg\min}} \, \mathcal{L}(f_{\text{tar}}(\theta_{\text{tar}}, x), y)}, \quad
    \text{s.t.}\, \theta_{\text{tar}} = \mathcal{T}\otimes \mathcal{S}_{\text{tar}}
\label{equ:train_S}
\end{equation}
Given the limited parameter count of the scalers (typically a few thousand), convergence is typically achieved within a few hundred iterations (around 0.16 epochs), making the computational cost negligible.
The model initialization process is completed once the training of weight scalers is finalized, after which subsequent training proceeds as usual without imposing additional constraints.

\section{Experiments}
\subsection{Benchmark for Evaluating Learngenes}
\textit{Learngene} offers an innovative approach to model initialization and knowledge transfer.
In this study, we propose a benchmark to comprehensively evaluate the efficiency of learngenes in initializing variable-sized models and their transferability across diverse datasets.

\textbf{Datasets.} 
We use ImageNet-1K~\cite{deng2009imagenet} as the primary dataset for knowledge integration and model evaluation. Additional datasets include Oxford Flowers, CUB-200-2011, Stanford Cars, CIFAR-10, CIFAR-100, Food-101, and iNaturalist-2019, providing a comprehensive benchmark to assess learngenes and other model initialization methods in terms of initialization flexibility, learning capacity, and knowledge transferability. Further details are provided in Appendix~\ref{app:dataset}.

\textbf{Network Structures.} 
We adopt DeiT~\cite{touvron2021training} as the base architecture and explore variations in both depth and width. 
For depth variations, we construct models with depths ranging from $D$ = 4, 6, 8, 10, to 12, with deeper configurations ($D$ = 24 and 36) provided in Appendix~\ref{app:deeper}.
For width variations, we adjust the number of attention heads while proportionally scaling the projection and MLP layers to target dimension ($W=$ 384, 576, 768, 1152, 1536). 
These configurations cover a wide range of ViT sizes, providing a thorough evaluation of learngene initialization.

\textbf{Measurements.} Learngene performance is primarily evaluated through Top-1 accuracy across various model sizes and datasets.
Additional considerations include extra training epochs and transferred parameters to highlight initialization efficiency.

\subsection{Experimental Setup}
\label{sec:setup}
To integrate knowledge from pre-trained models, we follow the basic settings of TLEG~\cite{xia2024transformer}, employing the pre-trained Levit-384~\cite{graham2021levit} and performing distillation for 150 epochs. Experiments with larger pre-trained models are provided in Appendix~\ref{app:larger_anc}.
In our experiments, we adopt DeiT-Ti, -S and -B (each with 12 layers) as auxiliary models and unify the sizes of weight templates across different components, including $T_{qkv}$, $T_{o}$, $T_{in}$ and $T_{out}$. 
Specifically, we set three basic sizes of weight templates, including 192$\times$192, 384$\times$384, and 768$\times$768, to accommodate basic scale variations (corresponding to DeiT-Ti, DeiT-S, and DeiT-B). Additional details are provided in Appendix~\ref{app:hyper} and~\ref{app:config_wt}.

\section{Results}
\renewcommand{\arraystretch}{0.81}
\begin{table*}
    \centering
    \setlength{\tabcolsep}{0.9 mm} 
    \caption{Performance of initializing models with variable depths on ImageNet-1K. ``Para.(M)'' represents the number of parameters for each model size (row) and the \textit{average} number of parameters transferred during model initialization (column).
    ``Epoch'' refers to additional epochs required for knowledge integration or network pre-training.
    ``---'' indicates failure to initialize the corresponding model size, while ``N/A'' denotes cases where a metric is not applicable. All models are trained for 10 epochs after initialization.}
    \vspace{-0.1in}
    \resizebox{\textwidth}{!}{
        \begin{tabular}{@{}llccccccc|cccccc|cccccc@{}}
        \toprule[1.5pt]
        & & & & \multicolumn{5}{c|}{$W_{\text{192}}$ (DeiT-Ti)} & & \multicolumn{5}{c|}{$W_{\text{384}}$ (DeiT-S)} & & \multicolumn{5}{c}{$W_{\text{768}}$ (DeiT-B)}\\
        \cmidrule{5-9}
        \cmidrule{11-15}
        \cmidrule{17-21}
        & & & & $L_4$ & $L_6$ & $L_8$ & $L_{10}$ & $L_{12}$ & & $L_4$ & $L_6$ & $L_8$ & $L_{10}$ & $L_{12}$ & & $L_4$ & $L_6$ & $L_8$ & $L_{10}$ & $L_{12}$ \\
        \cmidrule{5-9}
        \cmidrule{11-15}
        \cmidrule{17-21}
        \multicolumn{2}{l}{Methods} & Epoch & Para. & \cellcolor{gray!15}{2.2} & \cellcolor{gray!15}{3.1} & \cellcolor{gray!15}{4.0} & \cellcolor{gray!15}{4.9} & \cellcolor{gray!15}{5.8} 
            & Para. & \cellcolor{gray!15}{7.9} & \cellcolor{gray!15}{11.5} & \cellcolor{gray!15}{15.0} & \cellcolor{gray!15}{18.6} & \cellcolor{gray!15}{22.2} 
            & Para. & \cellcolor{gray!15}{29.9} & \cellcolor{gray!15}{44.1} & \cellcolor{gray!15}{58.3} & \cellcolor{gray!15}{72.5} & \cellcolor{gray!15}{86.7} \\
        \toprule[1.1pt]
        \multirow{3}{*}{\rotatebox{90}{\small{\textbf{Direct}}}}
        & He-Init~\cite{chen2021empirical} & N/A 
                & \cellcolor{gray!15}{0} & 34.7 & 40.6 & 43.7 & 46.8 & 48.3
                & \cellcolor{gray!15}{0} & 42.2 & 49.4 & 52.1 & 53.7 & 55.5
                & \cellcolor{gray!15}{0} & 47.9 & 53.1 & 54.4 & 55.0 & 56.7 \\
        & Mimetic~\cite{trockman2023mimetic} & N/A
                & \cellcolor{gray!15}{0} & 35.1 & 40.2 & 43.2 & 46.3 & 48.1
                & \cellcolor{gray!15}{0} & 43.3 & 49.1 & 53.0 & 54.1 & 55.6
                & \cellcolor{gray!15}{0} & 50.2 & 54.3 & 56.5 & 58.5 & 58.6 \\
        & GHN-3~\cite{knyazev2023canwescale} & N/A
                & \cellcolor{gray!15}{0} & 40.9 & 45.0 & 46.6 & 49.1 & 48.9
                & \cellcolor{gray!15}{0} & 45.4 & 49.0 & 50.2 & 52.3 & 53.2
                & \cellcolor{gray!15}{0} & 49.5 & 52.5 & 53.8 & 54.2 & 54.3 \\
        \midrule
        \multirow{2}{*}{\rotatebox{90}{\small{\textbf{Tran.}}}}
        & Share Init~\cite{lan2019albert} & 150
                   & \cellcolor{gray!15}{0.8} & 55.2 & 59.8 & 62.5 & 64.3 & 65.3
                   & \cellcolor{gray!15}{2.5} & 65.0 & 69.7 & 71.7 & 72.7 & 73.3
                   & \cellcolor{gray!15}{8.6} & 71.7 & 75.3 & 76.4 & 77.4 & 77.6 \\
        & LiGO~\cite{wang2023learning} & N/A
             & \cellcolor{gray!15}{2.2} & --- & 59.0 & 60.2 & 59.8 & 60.9
             & \cellcolor{gray!15}{7.9} & --- & 68.6 & 69.9 & 69.7 & 70.0
             & \cellcolor{gray!15}{29.9} & --- & 74.2 & 74.4 & 75.3 & 75.4 \\
        \midrule
        \multirow{4}{*}{\rotatebox{90}{\small{\textbf{Learngene}}}}
        & Heur-LG~\cite{wang2022learngene} & N/A
                & \cellcolor{gray!15}{1.7} & 41.5 & 47.4 & 50.5 & 53.5 & 55.5
                & \cellcolor{gray!15}{6.1} & 52.3 & 57.3 & 61.7 & 64.4 & 65.9
                & \cellcolor{gray!15}{22.8} & 60.5 & 68.7 & 72.2 & 73.6 & 74.0 \\
        & Auto-LG~\cite{wang2023learngene} & 50
                & \cellcolor{gray!15}{2.2} & 52.4 & 61.8 & 64.6 & 65.9 & 66.8
                & \cellcolor{gray!15}{7.9} & 63.2 & 70.5 & 72.2 & 73.3 & 73.8
                & \cellcolor{gray!15}{29.9} & 60.9 & 70.0 & 72.4 & 73.5 & 73.8 \\
        & TLEG~\cite{xia2024transformer} & 150
             & \cellcolor{gray!15}{1.3} & 55.0 & 60.5 & 62.9 & 64.4 & 65.4
             & \cellcolor{gray!15}{4.3} & 65.4 & 70.5 & 72.1 & 73.2 & 73.8
             & \cellcolor{gray!15}{15.7} & 71.6 & 74.9 & 76.2 & 77.0 & 77.1 \\
        & \cellcolor{blue!12}{WAVE} & \cellcolor{blue!12}{150}
               & \cellcolor{blue!12}{1.3} & \cellcolor{blue!12}{\textbf{58.6}} & \cellcolor{blue!12}{\textbf{63.2}} & \cellcolor{blue!12}{\textbf{65.4}} & \cellcolor{blue!12}{\textbf{66.6}} & \cellcolor{blue!12}{\textbf{67.3}}
               & \cellcolor{blue!12}{4.4} & \cellcolor{blue!12}{\textbf{68.9}} & \cellcolor{blue!12}{\textbf{72.7}} & \cellcolor{blue!12}{\textbf{74.1}} & \cellcolor{blue!12}{\textbf{74.9}} & \cellcolor{blue!12}{\textbf{75.3}}
               & \cellcolor{blue!12}{15.8} & \cellcolor{blue!12}{\textbf{74.5}} & \cellcolor{blue!12}{\textbf{77.5}} & \cellcolor{blue!12}{\textbf{78.2}} & \cellcolor{blue!12}{\textbf{78.9}} & \cellcolor{blue!12}{\textbf{79.2}}\\
               & & & & \textcolor{mygreen}{$\uparrow$3.4} 
                   & \textcolor{mygreen}{$\uparrow$1.4} 
                   & \textcolor{mygreen}{$\uparrow$0.8} 
                   & \textcolor{mygreen}{$\uparrow$0.7} 
                   & \textcolor{mygreen}{$\uparrow$0.5} 
                 & & \textcolor{mygreen}{$\uparrow$3.5} 
                   & \textcolor{mygreen}{$\uparrow$2.2} 
                   & \textcolor{mygreen}{$\uparrow$1.9} 
                   & \textcolor{mygreen}{$\uparrow$1.6} 
                   & \textcolor{mygreen}{$\uparrow$1.5} 
                 & & \textcolor{mygreen}{$\uparrow$2.8} 
                   & \textcolor{mygreen}{$\uparrow$2.2} 
                   & \textcolor{mygreen}{$\uparrow$1.8} 
                   & \textcolor{mygreen}{$\uparrow$1.5} 
                   & \textcolor{mygreen}{$\uparrow$1.6}\\
        \bottomrule[1.5pt]
        \end{tabular}   
        }
    \label{tab:lenth}
    \vspace{-0.07in}
\end{table*}

\subsection{Performance of Initializing Variable-sized Models}
In real-world scenarios, deployment constraints often necessitate models of varying sizes to meet specific needs. To demonstrate this, we construct 15 downstream models, each varying in depth and width, and compare WAVE against SOTA model initialization methods, categorized as follows: 
(1)~Direct Initialization: Methods like He-Init~\cite{chen2021empirical}, Mimetic Init~\cite{trockman2023mimetic}, and GHN-3~\cite{knyazev2023canwescale} use prior knowledge or hypernetworks to direct initialize models. 
(2)~Transfer Initialization: Techniques like Weight Selection~\cite{xu2023initializing}, LiGO~\cite{wang2023learning}, and Share Init~\cite{lan2019albert} transfer pre-trained knowledge to the target model.
(3)~Learngene Initialization: Methods such as Heur-LG~\cite{wang2022learngene}, Auto-LG~\cite{wang2023learngene}, and TLEG~\cite{xia2024transformer} use neural network fragments for model initialization.

\subsubsection{Depth Expansion}
Table~\ref{tab:lenth} presents the results of initializing models with variable depths (with deeper models detailed in Appendix~\ref{app:deeper}). Remarkably, WAVE demonstrates superior performance, significantly outperforming other learngene and model initialization methods across all model scales.

Transferring pre-trained knowledge for target initialization is more effective and straightforward than direct initialization, with Share Init and LiGO significantly outperforming Mimetic and GHN-3.
While LiGO enables smaller pre-trained models to initialize larger ones, it introduces numerous random parameters during weight transformation, highlighting challenges posed by size mismatches (see Table~\ref{tab:ana_kro}).
In contrast, Share Init improves performance by training and reusing specific layers, emphasizing the importance of rule-based knowledge integration before initialization.

However, existing learngene-based methods impose overly strict constraints, limiting learngenes to a few layers without reserving adaptive designs for diverse model sizes.
In contrast, WAVE treats each model's initialization as a distinct task, extracting a unified set of weight templates and enhancing flexibility through a few trainable parameters (i.e., weight scalers) to adapt connection rules to various model sizes.
This approach enables WAVE to integrate knowledge through predefined rules while allowing model-specific customization, achieving structural knowledge integration without excessive inductive bias.

\subsubsection{Width Expansion}
\begin{table}
    \centering
    \setlength{\tabcolsep}{0.9 mm}
    \vspace{-0.05in}
    \caption{Performance of initializing models with variable widths on ImageNet-1K. All models are trained for 10 epochs after initialization.}
    \vspace{-0.07in}
    \resizebox{0.48\textwidth}{!}{
        \begin{tabular}{@{}llccccccc@{}}
        \toprule[1.5pt]
        & & & & \multicolumn{5}{c}{$L_{\text{6}}$}\\
        \cmidrule{5-9}
        & & & & $W_{\text{384}}$ & $W_{\text{576}}$ & $W_{\text{768}}$ & $W_{\text{1152}}$ & $W_{\text{1536}}$ \\
        \cmidrule{5-9}
        \multicolumn{2}{l}{Methods} & Epoch & Para. & \cellcolor{gray!15}{11.5} & \cellcolor{gray!15}{25.1} & \cellcolor{gray!15}{44.1} & \cellcolor{gray!15}{98.0} & \cellcolor{gray!15}{173.1} \\ 
        \toprule[1.1pt]
        \multirow{3}{*}{\rotatebox{90}{\small{\textbf{Direct}}}}
        & He-Init~\cite{chen2021empirical} & N/A & \cellcolor{gray!15}{0} & 49.4 & 51.5 & 53.1 & 46.6 & 31.3 \\
        & Mimetic~\cite{trockman2023mimetic} & N/A & \cellcolor{gray!15}{0} & 49.1 & 48.0 & 54.3 & 47.7 & 33.0 \\ 
        & GHN-3~\cite{knyazev2023canwescale} & N/A & \cellcolor{gray!15}{0} & 49.0 & 51.8 & 52.5 & 52.7 & 51.0 \\
        \midrule
        \multirow{2}{*}{\rotatebox{90}{\small{\textbf{Tran.}}}}
        & Wt Select~\cite{xu2023initializing} & N/A & \cellcolor{gray!15}{26.8} &  48.6 & 50.7 & 55.4 & --- & ---  \\
        & LiGO~\cite{wang2023learning} & N/A & \cellcolor{gray!15}{10.0} & 63.6 & 63.1 & 69.5 & 70.0 & 73.7 \\
        
        \midrule
        \multirow{1}{*}{\rotatebox{90}{\small{\textbf{LG}}}}
        & \cellcolor{blue!12}{WAVE} & \cellcolor{blue!12}{150} & \cellcolor{blue!12}{5.4} & \cellcolor{blue!12}{\textbf{64.8}} & \cellcolor{blue!12}{\textbf{67.0}} & \cellcolor{blue!12}{\textbf{72.4}} & \cellcolor{blue!12}{\textbf{73.5}} & \cellcolor{blue!12}{\textbf{78.3}}  \\
        & & & & \textcolor{mygreen}{$\uparrow$1.2} 
                     & \textcolor{mygreen}{$\uparrow$3.9} 
                     & \textcolor{mygreen}{$\uparrow$2.9} 
                     & \textcolor{mygreen}{$\uparrow$3.5} 
                     & \textcolor{mygreen}{$\uparrow$4.6}\\
        \bottomrule[1.5pt]
        \end{tabular}
        }
    \label{tab:width}
    \vspace{-0.15in}
\end{table}

The adaptability of weight templates enables WAVE to scale the width of neural networks, achieving \textbf{\textit{what learngenes accomplish for the first time}} and consistently outperforming other model initialization and knowledge transfer methods, as shown in Table~\ref{tab:width}.

Similar to the observations in Table~\ref{tab:lenth}, methods like Mimetic and GHN-3 continue to underperform compared to parameter-transfer approaches, underscoring the importance of pre-trained knowledge for model initialization. 
LiGO persists in introducing excessive random parameters, while Weight Selection disrupts structural knowledge and fails to initialize models larger than the pre-trained ones.

By leveraging the Kronecker product, WAVE efficiently initializes models by concatenating weight templates with minimal parameters, making it the most effective approach for initializing models of varying widths.

\subsection{Compared with Direct Pre-Training}
\begin{figure}[tb]
  \centering
  \includegraphics[width=\linewidth]{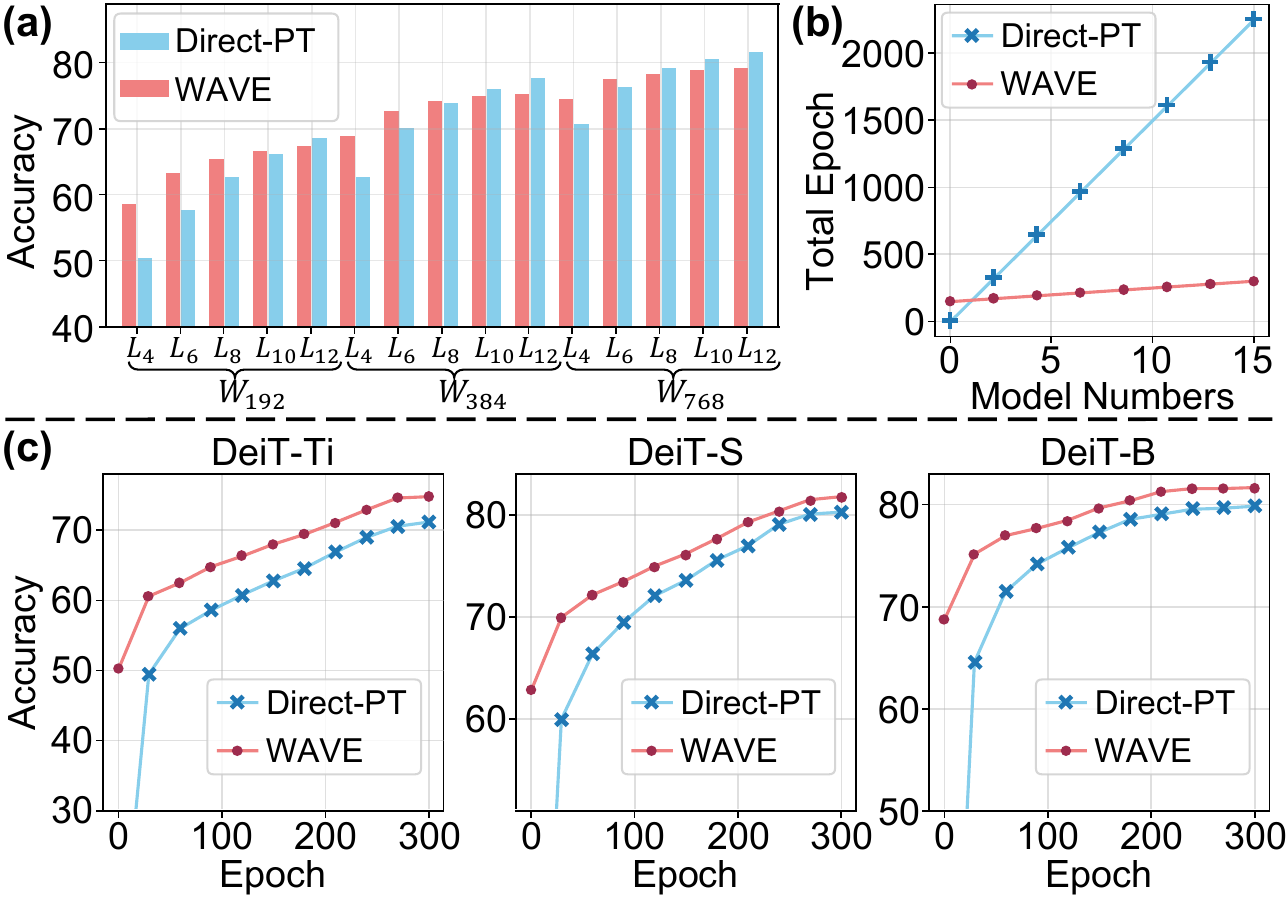}
  \vspace{-0.2in}
  \caption{Compared with Direct Pre-training. (a) Comparison of models initialized by WAVE and trained for 10 epochs versus those directly pre-trained for 150 epochs across 15 downstream models of varying sizes. (b) Analysis of computational cost as the number of initialized models increases.
  (c) Detailed training process (300 epochs) of models initialized by WAVE and direct pre-training.}
  \label{fig:pretrain}
  \vspace{-0.1in}
\end{figure}

\begin{table*}
    \centering 
    \setlength{\tabcolsep}{1mm}
    \caption{Performance of models on downstream datasets. ``Para.(M)'' is the average parameter transferred during model initialization.}
    \vspace{-0.05in}
    \resizebox{\textwidth}{!}{
        \begin{tabular}{@{}llcccccccc>{\columncolor{gray!15}}c|cccccccc>{\columncolor{gray!15}}c@{}}
        \toprule[1.5pt]
             & & \multicolumn{9}{c|}{DeiT-Ti, 3.0M} & \multicolumn{9}{c}{DeiT-S, 11.3M}\\
             \cmidrule{3-20}
             \multicolumn{2}{l}{Methods} & Para. & Flow. & CUB & Cars & C\small{10} & C\small{100} & Food & iNat. & \textit{Aver.}
             & Para. & Flow. & CUB & Cars & C\small{10} & C\small{100} & Food & iNat. & \textit{Aver.}\\
             \midrule[1.1pt]
             \multirow{3}{*}{\rotatebox{90}{\small{\textbf{Direct}}}}
             & He-Init~\cite{chen2021empirical} & 0 & 53.9 & 26.1 & 19.9 & 92.4 & 68.3 & 68.4 & 52.3 & \textit{54.5}
                     & 0 & 57.2 & 27.3 & 23.8 & 94.0 & 66.5 & 70.6 & 54.0 & \textit{56.2}\\
             & Mimetic~\cite{trockman2023mimetic} & 0 & 52.1 & 35.0 & 20.5 & 88.9 & 63.4 & 66.9 & 49.0 & \textit{53.7}
                          & 0 & 57.4 & 39.6 & 34.2 & 91.6 & 65.7 & 67.1 & 52.2 & \textit{58.3}\\ 
             & GHN-3~\cite{knyazev2023canwescale} & 0 & 50.0 & 41.1 & 23.2 & 92.5 & 70.1 & 76.2 & 51.7 & \textit{57.8}
                   & 0 & 52.7 & 45.2 & 30.6 & 93.9 & 72.7 & 76.2 & 55.5 & \textit{61.0}\\
             \midrule
             \multirow{3}{*}{\rotatebox{90}{\small{\textbf{Trans.}}}}
             & Wt Select~\cite{xu2023initializing} & 2.9 & 55.0 & 34.4 & 21.7 & 92.5 & 67.0 & 67.4 & 51.2 & \textit{55.6}
                           & 11.0 & 58.3 & 33.1 & 28.1 & 94.1 & 68.1 & 69.0 & 54.2 & \textit{57.8}\\
             & Share Init~\cite{lan2019albert} & 0.6 & 92.4 & 70.1 & 82.1 & 96.0 & 77.2 & 81.2 & 63.2 & \textit{80.3}
                        & 2.2 & 94.1 & 72.4 & 87.2 & 96.5 & 78.5 & 83.0 & 63.0 & \textit{82.1}\\
             & LiGO~\cite{wang2023learning} & 2.0 & 94.2 & 71.8 & 83.9 & 95.6 & 78.5 & 82.1 & 61.6 & \textit{81.1}
                        & 7.5 & 95.9 & 74.8 & 87.9 & 96.9 & 81.3 & 84.0 & 66.1 & \textit{83.8}\\
             \midrule
             \multirow{4}{*}{\rotatebox{90}{\small{\textbf{Learngene}}}}
             & Heur-LG~\cite{wang2022learngene} & 1.5 & 64.7 & 44.6 & 37.7 & 94.0 & 71.1 & 74.7 & 57.4 & \textit{63.5}
                     & 5.7 & 69.1 & 48.0 & 51.2 & 95.1 & 72.8 & 76.8 & 59.3 & \textit{67.5}\\
             & Auto-LG~\cite{wang2023learngene} & 2.0 & 93.5 & 71.4 & 83.5 & 96.4 & 77.1 & 81.7 & 62.5 & \textit{80.9}
                     & 7.5 & 96.4 & 75.1 & 88.2 & 97.3 & 81.0 & 84.6 & 67.0 & \textit{84.2}\\
             & TLEG~\cite{xia2024transformer} & 1.1 & 91.0 & 69.5 & 78.2 & 96.1 & 77.0 & 82.0 & 63.4 & \textit{79.6}
                  & 3.9 & 93.7 & 72.6 & 87.2 & 97.2 & 80.2 & 84.9 & 66.5 & \textit{83.2} \\
             & \cellcolor{blue!12}{WAVE} & \cellcolor{blue!12}{1.1} & \cellcolor{blue!12}{\textbf{94.9}} & \cellcolor{blue!12}{\textbf{74.8}} & \cellcolor{blue!12}{\textbf{84.4}} & \cellcolor{blue!12}{\textbf{96.6}} & \cellcolor{blue!12}{\textbf{80.7}} & \cellcolor{blue!12}{\textbf{83.8}} & \cellcolor{blue!12}{\textbf{65.2}} & \cellcolor{blue!12}{\textbf{\textit{82.9}}}
                  & \cellcolor{blue!12}{4.0} & \cellcolor{blue!12}{\textbf{96.9}} & \cellcolor{blue!12}{\textbf{78.1}} & \cellcolor{blue!12}{\textbf{89.4}} & \cellcolor{blue!12}{\textbf{97.4}} & \cellcolor{blue!12}{\textbf{83.2}} & \cellcolor{blue!12}{\textbf{85.5}} & \cellcolor{blue!12}{\textbf{67.6}} & \cellcolor{blue!12}{\textbf{\textit{85.4}}}\\
             & & & \textcolor{mygreen}{$\uparrow$0.7} & \textcolor{mygreen}{$\uparrow$3.0} & \textcolor{mygreen}{$\uparrow$0.5} & \textcolor{mygreen}{$\uparrow$0.2} & \textcolor{mygreen}{$\uparrow$2.2} & \textcolor{mygreen}{$\uparrow$1.7} & \textcolor{mygreen}{$\uparrow$1.8} & \textcolor{mygreen}{\textit{$\uparrow$1.8}}
                  & & \textcolor{mygreen}{$\uparrow$0.5} & \textcolor{mygreen}{$\uparrow$3.0} & \textcolor{mygreen}{$\uparrow$1.2} & \textcolor{mygreen}{$\uparrow$0.1} & \textcolor{mygreen}{$\uparrow$1.9} & \textcolor{mygreen}{$\uparrow$0.6} & \textcolor{mygreen}{$\uparrow$0.6} & \textcolor{mygreen}{\textit{$\uparrow$1.2}} \\
             \midrule
             \multirow{1}{*}{\rotatebox{90}{\small{\textbf{PT}}}} 
             & Direct FT & 2.9 & 95.4 & 75.1 & 86.5 & 96.6 & 80.2 & 84.0 & 66.9 & \textit{83.5}
                      & 11.0 & 96.4 & 77.0 & 89.4 & 97.5 & 82.8 & 85.6 & 69.3 & \textit{85.4}\\

             \bottomrule[1.5pt]
        \end{tabular}
    \label{tab:downstream}
    }
    \vspace{-0.07in}
\end{table*}

When target model sizes are incompatible with pre-trained ones, models initialized by WAVE achieve superior performance with significantly reduced training effort. As illustrated in Figure~\ref{fig:pretrain}c, models initialized by WAVE demonstrate superior performance from the very start of training. Notably, these models, trained for just 10 epochs, significantly outperform those directly pre-trained for 150 epochs (Figure~\ref{fig:pretrain}a), achieving a remarkable $15\times$ reduction in computational costs (Figure~\ref{fig:pretrain}b).
This advantage becomes particularly pronounced in smaller networks and scenarios with a larger number of initialized models, underscoring WAVE's effectiveness and efficiency as an initialization approach.

\subsection{Transferability of WAVE}
Table~\ref{tab:lenth} and Table~\ref{tab:width} have effectively demonstrated WAVE's initialization capability on ImageNet-1K.
Moreover, the knowledge encapsulated in weight templates is common enough for transfer across diverse downstream datasets. 
Table~\ref{tab:downstream} highlights WAVE's performance on various datasets.

Evidently, WAVE achieves significant improvements across diverse downstream datasets, highlighting its effectiveness in model initialization. In contrast, Mimetic and GHN-3 may underperform compared to He-Init on certain datasets, revealing limitations in the universality of prior knowledge and constraints on parameter adaptability.
Furthermore, methods like Weight Selection may experience negative transfer~\cite{rosenstein2005transfer}, emphasizing the importance of transferring core knowledge~\cite{feng2024transferring}.

Furthermore, the inadequacy of small datasets (e.g., Oxford Flowers, Stanford Car, and CUB-200-2011) to support large model training without effective knowledge transfer is evident.
This further underscores WAVE's ability to maximize data utilization, as only a small amount of data is required to leverage the structured knowledge in weight templates for adaptive model initialization.

\subsection{Ablation and Analysis}
\subsubsection{Effect of Initialization Rules on Concatenating Weight Templates}
For the rules when initializing models with weight templates, we compare the advantages of the Kronecker product with the \textit{linear combination} in TLEG~\cite{xia2024transformer} and the \textit{weight transformation} in LiGO~\cite{wang2023learning}.

As illustrated in Table~\ref{tab:ana_kro}, although linear combination avoids introducing additional trainable parameters, its predefined rules lack generality across different model sizes and do not support width expansion.
Conversely, weight transformation introduces numerous random parameters to expand pre-trained weight matrices, hindering the efficient adaptation of pre-trained knowledge within a limited training period.
In comparison, the Kronecker product offers a more flexible approach by concatenating weight templates with a small number of parameters, enabling efficient and adaptive initialization across various model sizes.

\begin{table}
    \centering
    \setlength{\tabcolsep}{1.5 mm}
    \caption{Ablation study on the Kronecker product. ``Init. Para.'' refers to the trainable parameters when initializing target models.}
    \vspace{-0.06in}
    \resizebox{0.474\textwidth}{!}{
        \begin{tabular}{@{}lccc|cc|cc@{}}
        \toprule[1.5pt]
        \multirow{6}{*}{\rotatebox{90}{\textbf{\makecell{Depth \\ Variation}}}} & & & & \multicolumn{2}{c|}{Linear} & \multicolumn{2}{c}{Kronecker}\\
        & $W$ & $L$ & FLOPs & Init. Para. & Acc. & Init. Para. & Acc.\\
        \cmidrule[1.1pt]{2-8}
        & \multirow{3}{*}{768} & 4 & 11.69 & 0 M & 71.6 & 0.0005 M & \cellcolor{blue!12}{\textbf{74.5}} \\
        & & 8 & 23.15 & 0 M & 76.2 & 0.0011 M & \cellcolor{blue!12}{\textbf{78.2}} \\
        & & 12 & 34.61 & 0 M & 77.1 & 0.0016 M & \cellcolor{blue!12}{\textbf{79.2}} \\
        \midrule[1.5 pt]
        \multirow{6}{*}{\rotatebox{90}{\textbf{\makecell{Width \\ Variation}}}} & & & & \multicolumn{2}{c|}{Transformation} & \multicolumn{2}{c}{Kronecker}\\
        & $W$ & $L$ & FLOPs & Init. Para. & Acc. & Init. Para. & Acc.\\
        \cmidrule[1.1pt]{2-8}
        & 384 & \multirow{3}{*}{6} & 4.55 & 4.28 M & 63.6 & 0.0025 M & \cellcolor{blue!12}{\textbf{64.8}} \\
        & 768 & & 17.42 & 17.11 M & 69.5 & 0.0025 M & \cellcolor{blue!12}{\textbf{72.4}} \\
        & 1536 & & 68.14 & 136.84 M & 73.7 & 0.0025 M & \cellcolor{blue!12}{\textbf{78.3}} \\
        \bottomrule[1.5pt]
        \end{tabular}
        }
    \label{tab:ana_kro}
    \vspace{-0.05in}
\end{table}

\subsubsection{Effect of WAVE on Different Components}
\begin{table}
    \centering
    \setlength{\tabcolsep}{2.5 mm}
    \caption{Ablation study on weight templates initializing different components of DeiT, with models consisting of 6 layers.}
    \vspace{-0.05in}
    \resizebox{0.474\textwidth}{!}{
        \begin{tabular}{@{}lccccccc@{}}
        \toprule[1.5pt]
        Methods & Att. & Proj. & FC & Norm & Ti & S & B\\
        \midrule[1.1pt]
        He-Init & & & & & 40.6 & 49.4 & 53.1 \\
        \midrule
        \multirow{5}{*}{WAVE} & & $\checkmark$ & &  & 50.2 & 57.9 & 60.4 \\
                                & $\checkmark$ & & & & 52.3 & 60.3 & 64.6 \\    
                                & & &$\checkmark$&  & 58.7 & 66.8 & 69.0 \\
                                & $\checkmark$ & $\checkmark$ & $\checkmark$ &  & 63.1 & 72.6 & 77.4 \\
                                & \cellcolor{blue!12}{$\checkmark$} & \cellcolor{blue!12}{$\checkmark$} & \cellcolor{blue!12}{$\checkmark$} & \cellcolor{blue!12}{$\checkmark$} & \cellcolor{blue!12}{\textbf{63.2}} & \cellcolor{blue!12}{\textbf{72.7}} & \cellcolor{blue!12}{\textbf{77.5}} \\
        \bottomrule[1.5pt]
        \end{tabular}
        }
    \label{tab:ablation}
    \vspace{-0.05in}
\end{table}
We further analyze the impact of weight templates on the initialization of different components in DeiT, as detailed in Table~\ref{tab:ablation}. 
The results indicate that all components can be effectively initialized using the structured knowledge encapsulated in weight templates.
In contrast, the weight matrices for normalization and bias are more data-dependent and have fewer parameters, which can be efficiently learned from the training data, making the use of weight templates for these components optional. 
The attention mechanism, consisting of Query, Key and Value, encodes more structured knowledge than other components, highlighting its critical role in the model's initialization~\cite{wang2024clusterlearngene}.

\subsubsection{Effect of Number and Shape of Templates}
\begin{table}
    \centering
    \setlength{\tabcolsep}{0.8 mm}
    \caption{Analysis of the number and shape of weight templates.}
    \vspace{-0.1in}
    \resizebox{0.48\textwidth}{!}{
        \begin{tabular}{@{}lccc|ccc|ccc@{}}
        \toprule[1.5pt]
        & \multicolumn{3}{c|}{$L_6$ $W_{192}$} & \multicolumn{3}{c|}{$L_6$ $W_{384}$} & \multicolumn{3}{c}{$L_6$ $W_{768}$}\\
        \cmidrule{2-10}
        & Para. & Shape & Acc. & Para. & Shape & Acc. & Para. & Shape & Acc. \\
        \cmidrule[1.1pt]{1-10}
        $\downarrow$ Num. & 0.9 & 192$^2$ & 57.5 & 
        2.6 & 384$^2$ & 68.4 & 
        8.6 & 768$^2$ & 75.7 \\        
        $\downarrow$ Shape & 1.3 & 96$^2$ & 60.3 & 
        4.4 & 192$^2$ & 71.6 & 
        15.8 & 384$^2$ & 75.8 \\
        \midrule
        \cellcolor{blue!12}{WAVE} & \cellcolor{blue!12}{1.3} & \cellcolor{blue!12}{192$^2$} & \cellcolor{blue!12}{\textbf{63.2}} & 
        \cellcolor{blue!12}{4.4} & \cellcolor{blue!12}{384$^2$} & \cellcolor{blue!12}{\textbf{72.7}} & 
        \cellcolor{blue!12}{15.8} & \cellcolor{blue!12}{768$^2$} & \cellcolor{blue!12}{\textbf{77.5}} \\
        \bottomrule[1.5pt]
        \end{tabular}
        }
    \label{tab:analysis}
\end{table}
We analyze the influence of the number and size of weight templates on initializing downstream models. 
As shown in Table~\ref{tab:analysis}, reducing the number of weight templates significantly degrades model performance, as fewer parameters fail to capture the full spectrum of size-agnostic knowledge.
Additionally, we experiment with reducing the size of weight templates to enhance flexibility in width expansion. 
However, such flexibility sacrifices the performance for the insufficient structured knowledge in weight templates.

Therefore, both the number and size of weight templates are crucial for effective initialization.
WAVE achieves an optimal balance between performance and cost by utilizing approximately 16.7\% of the auxiliary model's parameters.
The basic size of weight templates is aligned with the \textit{embedding dimensions} of the auxiliary model (refer to Section~\ref{sec:setup} for details) to facilitate the effective capture of structured knowledge, as illustrated in Figure~\ref{fig:structure}a.

\subsection{Visualization of Knowledge in Weight Templates}
\subsubsection{Visualization of Structured Knowledge}
\begin{figure}[tb]
  \centering
  \includegraphics[width=\linewidth]{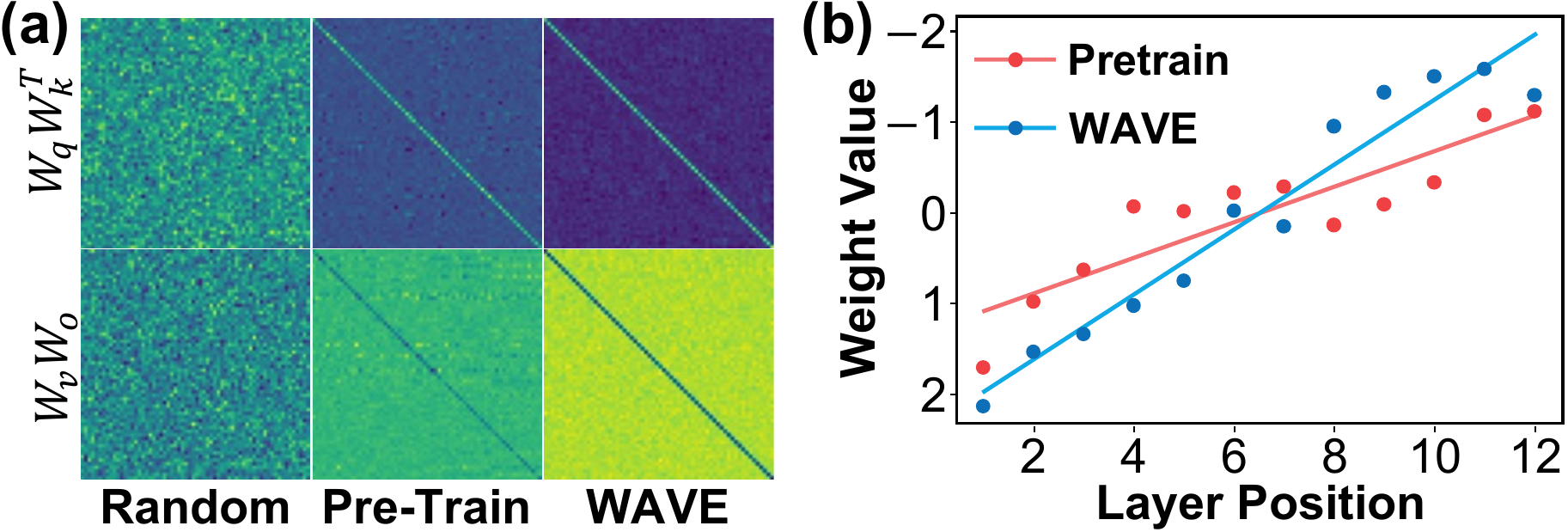}
  \vspace{-0.2in}
  \caption{Visualization of structured knowledge. (a) Knowledge in self-attention layers. (b) Relationships between layer position and corresponding parameter values after PCA.}
  \label{fig:structure}
  \vspace{-0.1in}
\end{figure}

Mimetic Initialization~\cite{trockman2023mimetic} identifies diagonal properties in self-attention layers, while TLEG~\cite{xia2024transformer} captures linear correlations across layers. 
However, these observations are exclusive to pre-trained ViTs, with both Mimetic Initialization and TLEG manually preserving such structured knowledge during initialization.
Remarkably, as demonstrated in Figure~\ref{fig:structure}, WAVE autonomously encapsulates such structured knowledge from pre-trained models into weight templates without manual intervention.
Consequently, models initialized by weight templates inherently retain these structural characteristics in their parameter matrices.

\subsubsection{Visualization of Common Knowledge}
\begin{figure}[tb]
  \centering
  \includegraphics[width=\linewidth]{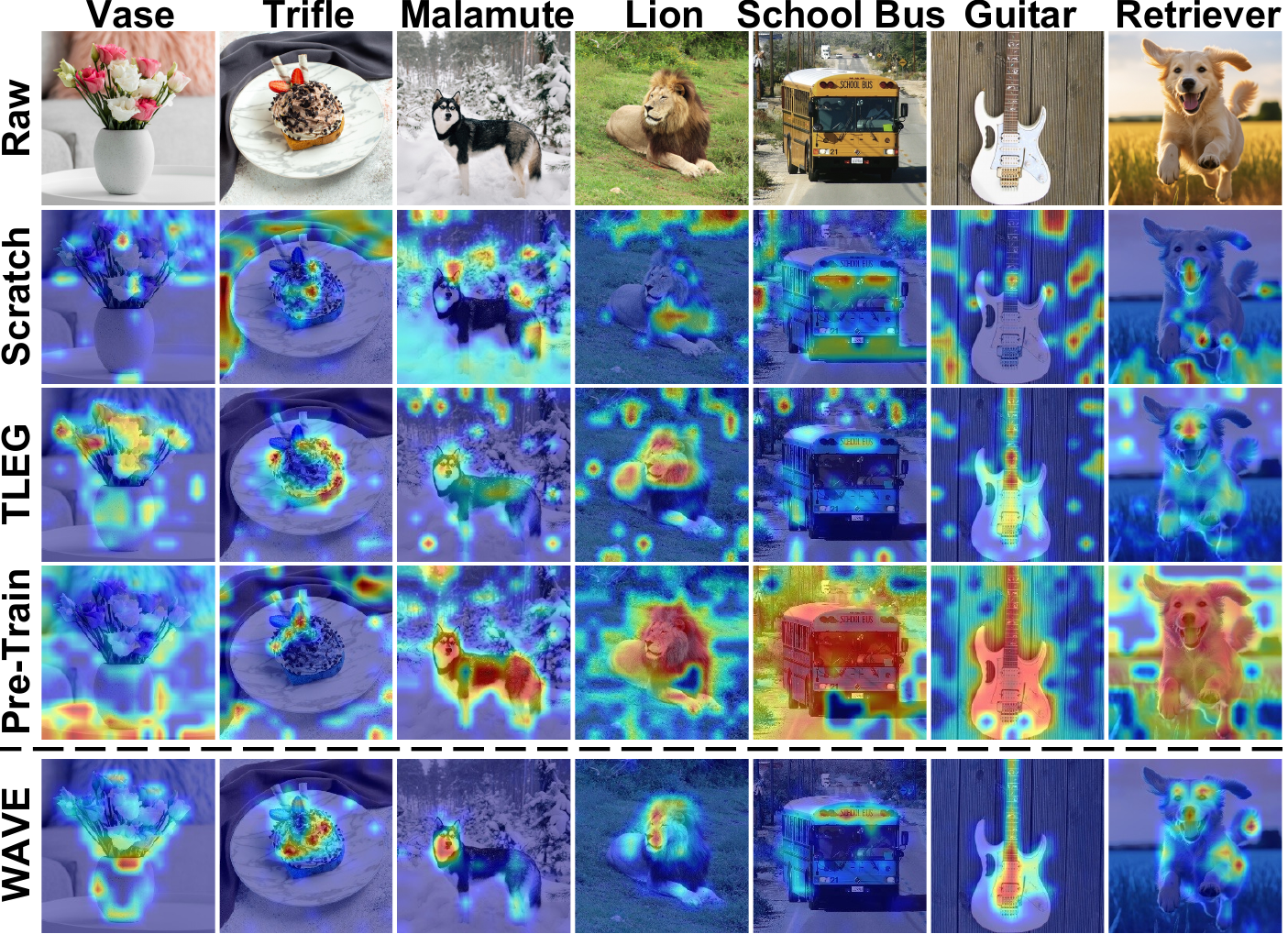}
  \vspace{-0.25in}
  \caption{Visualization of knowledge encapsulated in weight templates. All networks operate directly after initialization without any additional training or fine-tuning.}
  \label{fig:visual}
  \vspace{-0.1in}
\end{figure}
In addition, we demonstrate that weight templates enable neural networks to focus more on the common local features after initialization. 
We select sample images and employ CAM~\cite{selvaraju2017grad} to visualize the attention in pre-trained networks, as well as those initialized by learngenes (e.g., TLEG).

As shown in Figure~\ref{fig:visual}, random initialization shows scattered and widespread attention, whereas pre-trained models transfer an entire set of previously learned knowledge, leading to broader attention that often includes irrelevant details, such as the image background.
Both TLEG and WAVE focus more on local features (i.e., smaller red attention blocks), with WAVE demonstrating superior localization and a cleaner focus (i.e., removing attention on image background), thereby enhancing classification performance.

\section{Conclusion}
In this paper, we introduce WAVE, an innovative approach for initializing models of variable sizes. 
WAVE leverages shared weight templates that encapsulate size-agnostic knowledge, along with size-specific weight scalers, to effectively extract and transfer structured knowledge from pre-trained models for initialization across all model sizes. 
WAVE demonstrates superior performance in initializing models for both depth and width expansions, and the size-agnostic knowledge is also common enough to be transferred to various downstream datasets. 

\section*{Acknowledgement}
We sincerely thank Wenqian Li, Jianlu Shen, and Ruixiao Shi for their insightful discussions on this work. We also appreciate Freepik for contributing to the figure design. 
This research is supported by the Science and Technology Major Project of Jiangsu Province under Grant BG20240305, Key Program of Jiangsu Science Foundation under Grant BK20243012, National Natural Science Foundation of China under Grants U24A20324, 62125602, and 62306073, Natural Science Foundation of Jiangsu Province under Grant BK20230832, and the Xplorer Prize.

{
    \small
    \bibliographystyle{ieeenat_fullname}
    \bibliography{main}
}

\clearpage
\appendix
\counterwithin{figure}{section}
\counterwithin{table}{section}
\counterwithin{equation}{section}

\section{Universality of WAVE}
\label{sec:proof}
\textbf{Proposition 1.} \textit{Heur-LG~\cite{wang2022learngene}, Auto-LG~\cite{wang2023learngene} and TLEG~\cite{xia2024transformer} are special cases of WAVE.}\\
\textbf{Proof.} To prove Proposition 1, we establish a correspondence between weight templates and ViT layers since the learngenes in Heur-LG, Auto-LG and TLEG are structured in the form of ViT layers. 

Consider the set of weight templates:
\begin{equation*}
    \mathcal{T} = \{T_{qkv}^{(1\thicksim N_{qkv})}, T_{o}^{(1\thicksim N_o)}, T_{in}^{(1\thicksim N_{in})}, T_{out}^{(1\thicksim N_{out}})\}
\end{equation*}

These templates can be used to construct $N_l$ learngene layers in a ViT model:
\begin{equation*}
    \mathcal{G} = \{G_{qkv}^{(1\thicksim N_l)}, G_{o}^{(1\thicksim N_l)}, G_{in}^{(1\thicksim N_l)}, G_{out}^{(1\thicksim N_l)}\}
\end{equation*}
via the following operation:
\begin{equation}
    G_{\star}^{(l)} = \sum_{t=1}^{s_1 s_2}T_{\star}^{(s_1 s_2 \cdot (l-1) + t)} \otimes \mathds{\mathring{1}}_{(i, j)}
\end{equation}
where $G_{\star}^{(l)}\in \mathbb{R}^{M_1\times M_2}$ is the constructed weight matrix for the $l$-th layer of type $\star$. $T_{\star}^{(t)}\in \mathbb{R}^{w_1\times w_2}$ is the $t$-th weight template of type $\star$, where $\star \in \{qkv, o, in, out\}$. $\mathds{\mathring{1}}_{(i, j)}\in \mathbb{R}^{s_1\times s_2}$ is a padding matrix with 1 at position $(i, j)$ and 0 elsewhere, where $s_1 = \frac{M_1}{w_1}$ and $s_2 = \frac{M_2}{w_2}$.

The indices $i$ and $j$ in $\mathds{\mathring{1}}_{(i, j)}$ are calculated as follows:
\begin{equation}
    i=\lfloor \frac{t-1}{s_2}\rfloor,\;\; j=(t-1)\bmod s_2
\end{equation}

Now, consider a ViT model with $L_{\text{tar}}$ layers whose weight matrices are 
\begin{equation*}
    \mathcal{W} = \{W_{qkv}^{(1\thicksim L_{\text{tar}})}, W_{o}^{(1\thicksim L_{\text{tar}})}, W_{in}^{(1\thicksim L_{\text{tar}})}, W_{out}^{(1\thicksim L_{\text{tar}})}\}
\end{equation*}

We will demonstrate how Heur-LG, Auto-LG, and TLEG can be represented as special cases of WAVE:
\begin{itemize}
    \item \textbf{Heur-LG} extracts the last $N_l$ layers from a pre-trained model and then stacks randomly initialized layers $R_{\star}$ in the lower layers to construct target models:
    \begin{equation}
        W_{\star}^{(l)} = 
            \begin{cases} 
                R_{\star}^{(l)} & l < L_{\text{tar}} - N_l \\ 
                G_{\star}^{(N_l+l-L_{\text{tar}})} & l \geq L_{\text{tar}} - N_l 
            \end{cases}
    \end{equation}
    
    \item \textbf{Auto-LG} extracts the first $N_l$ layers from a pre-trained model and then stacks randomly initialized layers $R_{\star}$ in the higher layers to construct target models:
    \begin{equation}
        W_{\star}^{(l)} = 
            \begin{cases} 
                G_{\star}^{(l)} & l \leq N_l \\ 
                R_{\star}^{(l)} & l > N_l 
            \end{cases}
    \end{equation}
    
    \item \textbf{TLEG} adopts linear expansion on two shared learngene layers $G_{\star}^{(\mathcal{A})}$ and $G_{\star}^{(\mathcal{B})}$:
    \begin{equation}
        W_{\star}^{(l)} = G_{\star}^{(\mathcal{A})} + \frac{l}{L_{\text{tar}}}G_{\star}^{(\mathcal{B})}
    \end{equation}
\end{itemize}

The above formulations demonstrate that Heur-LG, Auto-LG, and TLEG are all specific cases of WAVE, where different rules are applied to concatenate and weight these templates in $\mathcal{T}$.

\renewcommand{\arraystretch}{1}
\begin{algorithm}[tb]
    \caption{Integration of Structured Size-agnostic Knowledge into Weight Templates}
    \small
    \label{alg:algorithm}
    \textbf{Input}: Training dataset $\{(x^{(i)}, y^{(i)})\}_{i=1}^m$, number of epochs $N_{\text{ep}}$, batch size $B$, learning rate $\eta$, pre-trained model (i.e., ancestry model) $f_{\text{pre}}$ with weight matrices $\mathcal{W}$\\
    \textbf{Output}: Weight Templates $\mathcal{T}$
    \begin{algorithmic}[1]
        \STATE Random initialize Weight Matrices  $\mathcal{W}$, Weight Templates $\mathcal{T}$ and Weight Scalers $\mathcal{S}$
        \FOR{$ep = 1$ to $N_{\text{ep}}$}
            \FOR{each batch $\{(x_i, y_i)\}_{i=1}^B$}
                \STATE Update $\mathcal{W}$ with $\mathcal{T}$ and $\mathcal{S}$ under the rule of Eq.~\eqref{equ:kro}
                \STATE For each $x_i$, forward propagate $\hat{y}_i = f_{\text{aux}}(x_i)$
                \STATE Calculate $\mathcal{L}_{\text{batch}}=\frac{1}{B} \sum_{i=1}^B \mathcal{L}(\hat{y}_i, y_i)$
                \STATE Backward propagate $\mathcal{L}(\hat{y}_i, y_i)$ to compute the gradients with respect to $\mathcal{T}$ and $\mathcal{S}$: $\nabla_\mathcal{T} \mathcal{L}_{\text{batch}}, \nabla_\mathcal{S} \mathcal{L}_{\text{batch}}$
                \STATE Update $\mathcal{T}$ and $\mathcal{S}:$\\ \quad $\mathcal{T} := \mathcal{T} - \eta \cdot \nabla_\mathcal{T} \mathcal{L}_{\text{batch}}$\\ \quad
                $\mathcal{S} := \mathcal{S} - \eta \cdot \nabla_\mathcal{S} \mathcal{L}_{\text{batch}}$
            \ENDFOR
        \ENDFOR
    \end{algorithmic}
\end{algorithm}

\section{Initialization of Weight Scalers}
\label{app:init_ws}
For the initialization of $\mathcal{S}$, we simulate linear initialization~\cite{xia2024transformer} and Net2Net~\cite{chen2015net2net}, and propose \textbf{linear padding initialization} to better preserve the structured knowledge of original weight templates during initialization, thereby providing a suitable starting point for target networks. 

For a target network with $L_\text{tar}$ layers, we consider its weight matrix $W_{\star}^{(l)}\in \mathbb{R}^{M_1\times M_2}$ and corresponding weight templates $T_{\star}^{(1\thicksim N_{\star})}\in \mathbb{R}^{w_1\times w_2}$, where $M_1>w_1$ and $M_2>w_2$. The corresponding $S_{\star}^{(l, t)}\in \mathbb{R}^{s_1\times s_2}$ is initialized as:
\begin{equation}
    S_{\star}^{(l, t)} = \alpha_t \cdot \mathds{\mathring{1}}_{(i, j)} + \epsilon \mathcal{N}(\mu, \sigma^2)
\label{equ:init_S}
\end{equation}
Here, $\alpha_t = \begin{cases} 1, & \text{if } \small{t \le \frac{N_\star}{2}} \\\frac{l}{L_\text{tar}}, & \text{otherwise} \end{cases}$ 
is a linear weight and $\mathds{\mathring{1}}_{(i, j)}\in \mathbb{R}^{s_1\times s_2}$ is a padding matrix with 1 at $(i, j)$ and 0 elsewhere. The indices $(i,j)$ are given by $i=\lfloor \frac{(t-1) \bmod (s_1\times s_2)}{s_2}\rfloor$, $j=(t-1)\bmod s_2$ with $s_i = \frac{M_i}{w_i}$.
$\epsilon$ denotes a small value (e.g., 10$^{-6}$) and $\mathcal{N}(\mu, \sigma^2)$ represents Gaussian noise.

\section{Training Details}
\subsection{Details of Knowledge Integration}
\label{app:condense}
Algorithm \ref{alg:algorithm} presents the pseudo code for integrating structured size-agnostic knowledge into weight templates (i.e., learngenes).

\subsection{Hyper-parameters}
\label{app:hyper}
Table~\ref{tab:hyper_main} and Table~\ref{tab:hyper_down} present the basic settings, including batch size, warmup epochs, training epochs and other settings for WAVE integrating structured common knowledge into weight templates and training the models initialized with weight templates on various datasets, respectively.

\subsection{Details of Weight Templates}
\label{app:config_wt}
Table \ref{tab:config_wt} presents the details of weight templates used for integrating structured size-agnostic knowledge from the ancestry model of DeiT-Ti, DeiT-S and DeiT-B. 

\subsection{Details of Downstream Datasets}
\label{app:dataset}
Additional datasets include Oxford Flowers~\cite{nilsback2008automated}, CUB-200-2011~\cite{wah2011caltech}, Stanford Cars~\cite{gebru2017fine}, CIFAR-10, CIFAR-100~\cite{krizhevsky09}, Food-101~\cite{bossard2014food}, and iNaturalist-2019~\cite{tan2019herbarium}.
Table~\ref{tab:datasets} presents the details of seven downstream datasets, which are sorted by the size of datasets.

\renewcommand{\arraystretch}{0.9}
\begin{table}
    \centering
    \caption{Hyper-parameters for WAVE integrating structured knowledge on ImageNet-1K.}
    \vspace{-0.1in}
    \setlength{\tabcolsep}{4.6 mm}
        \begin{tabular}{@{}lr@{}}
        \toprule[1.3pt]
        \textbf{Training Settings} & \textbf{Configuration} \\
        \midrule[1.1pt]
        optimizer & AdamW\\
        base learning rate & Ti: 5e-4 $\mid$ S: 2.5e-4 $\mid$ B: 1.25e-4\\
        warmup learning rate & 1e-6\\
        weight decay & 0.05\\
        optimizer momentum & 0.9\\
        batch size & Ti: 512 $\mid$ S: 256 $\mid$ B: 128\\
        training epochs & 150\\
        learning rate schedule & cosine decay\\
        warmup epochs & 5\\
        color jitter & 0.4 \\
        auto augment & rand-m9-mstd0.5-inc1\\
        mixup & 0.8\\
        cutmix & 1.0\\
        label smoothing & 0.1\\
        drop path & 0.1\\
        \bottomrule[1.3pt]
        \end{tabular}
    \label{tab:hyper_main}
    \vspace{-0.05in}
\end{table}

\begin{table}
    \centering
    \setlength{\tabcolsep}{3mm}
    \caption{Configuration of weight templates. $l\times w$ @ $n$ represents that the weight templates of corresponding weight matrices are composed of $n$ templates with the size $l \times w$.}
    \vspace{-0.1in}
    \resizebox{0.48\textwidth}{!}{
        \begin{tabular}{@{}lccc@{}}
        \toprule[1.3pt]
         & \textbf{DeiT-Ti} & \textbf{DeiT-S} & \textbf{DeiT-B}\\
        \cmidrule[1.1pt]{1-4}
        $W_{qkv}$ & 192$\times$192 @ 6 & 384$\times$384 @ 6 & 768$\times$768 @ 6 \\
        $W_{o}$ & 192$\times$192 @ 2 & 384$\times$384 @ 2 & 768$\times$768 @ 2\\
        $W_{in}$ & 192$\times$192 @ 8 & 384$\times$384 @ 8 & 768$\times$768 @ 8\\
        $W_{out}$ & 192$\times$192 @ 8 & 384$\times$384 @ 8 & 768$\times$768 @ 8\\
        $W_{norm_1}$ & 192 @ 4 & 384 @ 4 & 768 @ 4\\
        $W_{norm_2}$ & 192 @ 4 & 384 @ 4 & 768 @ 4\\
        $W_{qkv}^{(\text{bias})}$ & 576 @ 4 & 1152 @ 4 & 2304 @ 4\\
        $W_{o}^{(\text{bias})}$ & 192 @ 4 & 384 @ 4 & 768 @ 4\\
        $W_{in}^{(\text{bias})}$ & 768 @ 4 & 1536 @ 4 & 3702 @ 4\\
        $W_{out}^{(\text{bias})}$ & 192 @ 4 & 384 @ 4 & 768 @ 4\\
        $W_{norm_1}^{(\text{bias})}$ & 192 @ 4 & 384 @ 4 & 768 @ 4\\
        $W_{norm_2}^{(\text{bias})}$ & 192 @ 4 & 384 @ 4 & 768 @ 4\\
        $W_{irpe_q}$ & 1$\times$64$\times$49 @ 6 & 1$\times$64$\times$49 @ 6 & 1$\times$64$\times$49 @ 6\\
        $W_{irpe_k}$ & 1$\times$64$\times$49 @ 6 & 1$\times$64$\times$49 @ 6 & 1$\times$64$\times$49 @ 6\\
        $W_{irpe_v}$ & 1$\times$49$\times$64 @ 6 & 1$\times$49$\times$64 @ 6 & 1$\times$49$\times$64 @ 6\\
        \bottomrule[1.3pt]
        \end{tabular}
        }
    \vspace{-0.1in}
    \label{tab:config_wt}
\end{table}

\begin{table}
    \centering
    \setlength{\tabcolsep}{1.3 mm}
    \caption{Characteristics of downstream datasets.}
    \vspace{-0.1in}
    \resizebox{0.48\textwidth}{!}{
        \begin{tabular}{@{}lcccc@{}}
        \toprule[1.3pt]
        \textbf{Dataset} & \textbf{Classes} & \textbf{Total} & \textbf{Training} & \textbf{Testing} \\
        \cmidrule[1.1pt]{1-5}
        \textbf{Oxford Flowers}~\cite{nilsback2008automated} & 102 & 8,189 & 2,040  & 6,149\\
        \textbf{CUB-200-2011}~\cite{wah2011caltech} & 200 & 11,788 & 5,994 & 5,794 \\
        \textbf{Stanford Cars}~\cite{gebru2017fine} & 196 & 16,185 & 8,144 & 8,041\\
        \textbf{CIFAR10}~\cite{krizhevsky09} & 10 & 60,000 & 50,000 & 10,000 \\
        \textbf{CIFAR100}~\cite{krizhevsky09} & 100 & 60,000 & 50,000 & 10,000 \\
        \textbf{Food101}~\cite{bossard2014food} & 101 & 101,000 & 75,750 & 25,250\\
        \textbf{iNat-2019}~\cite{tan2019herbarium} & 1010 & 268,243 & 265,213 & 3,030\\
        \bottomrule[1.3pt]
        \end{tabular}
        }
    \label{tab:datasets}
\end{table}

\begin{table*}
    \centering
    \setlength{\tabcolsep}{0.7 mm}
    \caption{Hyper-parameters for neural networks trained on downstream datasets.}
    \vspace{-0.1in}
    \resizebox{\textwidth}{!}{
        \begin{tabular}{@{}lccccccccccccc@{}}
        \toprule[1.3pt]
        \textbf{Dataset} & \makecell{\textbf{Batch}\\ \textbf{Size}} & \makecell{\textbf{Epoch}} & \makecell{\textbf{Learning}\\ \textbf{Rate}} & \makecell{\textbf{Drop}\\ \textbf{Last}} & \makecell{\textbf{Warmup}\\ \textbf{Epochs}} & \makecell{\textbf{Droppath}\\ \textbf{Rate}} & \makecell{\textbf{Color}\\ \textbf{Jitter}} & \makecell{\textbf{Auto} \\ \textbf{Augment}} & \makecell{\textbf{Random}\\ \textbf{Rrase}} & \makecell{\textbf{Mixup}} & \makecell{\textbf{Cutmix}} & \makecell{\textbf{Scheduler}} & \makecell{\textbf{Optimizer}}\\
        \midrule[1.1pt]
        \textbf{Oxford Flowers} & 512 & 300 & 3e-4 & False & 0 & 0 & 0.4 & \multirow{7}{*}{\rotatebox{90}{\fontsize{8.5}{12}\selectfont rand-m9-mstd0.5-inc1}} & 0.25 & 0 & 0 & cosine & AdamW \\
        \textbf{CUB-200-2011} & 512 & 300 & 3e-4 & False & 0 & 0.1 & 0 & & 0.25 & 0 & 0 & cosine & AdamW \\
        \textbf{Stanford Cars} & 512 & 300 & 3e-4 & False & 0 & 0.1 & 0 & & 0.25 & 0 & 0 & cosine & AdamW \\
        \textbf{CIFAR10} & 512 & 300 & 5e-4 & True & 0 & 0.1 & 0.4 & & 0.25 & 0 & 0 & cosine & AdamW\\
        \textbf{CIFAR100} & 512 & 300 & 5e-4 & True & 0 & 0.1 & 0.4 & & 0.25 & 0 & 0 & cosine & AdamW\\
        \textbf{Food101} & 512 & 300 & 5e-4 & True & 0 & 0.1 & 0.4 & & 0.25 & 0 & 0 & cosine & AdamW\\
        \textbf{iNat-2019} & 512 & 100 & 5e-4 & True & 0 & 0.1 & 0.4 & & 0.25 & 0 & 0 & cosine & AdamW\\
        \bottomrule[1.3pt]
        \end{tabular}
        }
    \label{tab:hyper_down}
\end{table*}

\section{Additional Results}
\begin{table}[h]
    \centering
    \setlength{\tabcolsep}{0.4 mm}
    \caption{Additional results on different ancestry models.}
    \vspace{-0.1in}
    \resizebox{0.48\textwidth}{!}{
        \begin{tabular}{@{}lclll|lll@{}}
        \toprule[1.3pt]
        & & \multicolumn{3}{c}{DeiT-Ti} & \multicolumn{3}{c}{DeiT-S}\\
        \cmidrule{3-8}
        & Ancestry & \multicolumn{1}{c}{$L_{\text{4}}$} & \multicolumn{1}{c}{$L_{\text{8}}$} & \multicolumn{1}{c}{$L_{\text{12}}$} & \multicolumn{1}{c}{$L_{\text{4}}$} & \multicolumn{1}{c}{$L_{\text{8}}$} & \multicolumn{1}{c}{$L_{\text{12}}$} \\
        \toprule[1.1pt]
        TLEG~\cite{xia2024transformer} & LeVit-384 (39.1M) & 55.0 & 62.9 & 65.4 & 65.4 & 72.1 & 73.8 \\
        \midrule
        \cellcolor{blue!12}{WAVE} & \cellcolor{blue!12}{LeVit-384 (39.1M)} & \cellcolor{blue!12}{58.6$^*$} & \cellcolor{blue!12}{65.4$^*$} & \cellcolor{blue!12}{\textbf{67.3}} & \cellcolor{blue!12}{\textbf{68.9}} & \cellcolor{blue!12}{\textbf{74.1}} & \cellcolor{blue!12}{75.3$^*$} \\
        \cellcolor{red!10}{WAVE} & \cellcolor{red!10}{RegNet-16GF (83.6M)} & \cellcolor{red!10}{\textbf{58.7}} & \cellcolor{red!10}{\textbf{65.7}} & \cellcolor{red!10}{67.0$^*$} & \cellcolor{red!10}{68.7$^*$} & \cellcolor{red!10}{74.0$^*$} & \cellcolor{red!10}{\textbf{75.4}} \\
        \bottomrule[1.3pt]
        \end{tabular}
        }
    \label{tab:anc}
\end{table}

\begin{table}[h]
    \centering
    \setlength{\tabcolsep}{0.6 mm}
    \caption{Results on initializing deeper models.}
    \vspace{-0.1in}
    \resizebox{0.48\textwidth}{!}{
        \begin{tabular}{@{}lcccc|ccc|ccc@{}}
        \toprule[1.5pt]
        & & & \multicolumn{2}{c|}{$W_{\text{192}}$} & & \multicolumn{2}{c|}{$W_{\text{384}}$} & & \multicolumn{2}{c}{$W_{\text{768}}$}\\
        \cmidrule{4-5}
        \cmidrule{7-8}
        \cmidrule{10-11}
        & & & $L_{\text{24}}$ & $L_{\text{36}}$ & & $L_{\text{24}}$ & $L_{\text{36}}$ & & $L_{\text{24}}$ & $L_{\text{36}}$\\
        \cmidrule{4-5}
        \cmidrule{7-8}
        \cmidrule{10-11}
        & Epoch & Para. & \cellcolor{gray!15}{11.3} & \cellcolor{gray!15}{16.7} & Para. & \cellcolor{gray!15}{43.6} & \cellcolor{gray!15}{65.0} & Para. & \cellcolor{gray!15}{171.8} & \cellcolor{gray!15}{257.0} \\
        \midrule[1.1pt]
        He-Init~\cite{chen2021empirical} & 0 & \cellcolor{gray!15}{0} & 52.4 & 53.6 & \cellcolor{gray!15}{0} & 57.8 & 58.2 & \cellcolor{gray!15}{0} & 59.2 & 59.4 \\
        TLEG~\cite{xia2024transformer} & 150 & \cellcolor{gray!15}{1.3} & 68.1 & 68.7 & \cellcolor{gray!15}{4.3} & 74.8 & 75.3 & \cellcolor{gray!15}{15.7} & 77.6 & 77.5 \\
        \midrule
        \cellcolor{blue!12}{WAVE} & \cellcolor{blue!12}{150} & \cellcolor{blue!12}{1.3}
        & \cellcolor{blue!12}{\textbf{68.9}} & \cellcolor{blue!12}{\textbf{69.5}} & \cellcolor{blue!12}{4.4} & \cellcolor{blue!12}{\textbf{75.5}} & \cellcolor{blue!12}{\textbf{76.6}} & \cellcolor{blue!12}{15.8} & \cellcolor{blue!12}{\textbf{79.4}} & \cellcolor{blue!12}{\textbf{79.6}} \\
        & & & \textcolor{mygreen}{$\uparrow$0.8} & \textcolor{mygreen}{$\uparrow$0.8} & & \textcolor{mygreen}{$\uparrow$0.7} & \textcolor{mygreen}{$\uparrow$1.3} & & \textcolor{mygreen}{$\uparrow$1.8} & \textcolor{mygreen}{$\uparrow$2.1}\\
        \bottomrule[1.5pt]
        \end{tabular}
        }
    \label{tab:deeper}
\end{table}

\subsection{Integration of Knowledge from Larger Pre-trained Models}
\label{app:larger_anc}
Weight templates enable structured integration of knowledge from pre-trained ancestor models while filtering out size-specific information that violates the constraints in Eq.~\eqref{equ:kro}. 
This mechanism ensures effective transfer and sharing of size-agnostic knowledge across models of varying sizes.

To evaluate the influence of ancestor models with different architectures and sizes, we incorporate a larger pre-trained model, RegNet-16GF (83.6M)~\cite{radosavovic2020designing}, and compare it with WAVE and TLEG, both using LeVit-384 (39.1M) as the ancestor model. Table~\ref{tab:anc} presents the results on DeiT-Ti and DeiT-S across various layers.

The results demonstrate that WAVE consistently outperforms TLEG across all model sizes. 
While the larger ancestor model (RegNet-16GF) provides some improvements, the gains remain limited, suggesting that once a pre-trained model is sufficiently trained and informative, the shared size-agnostic knowledge remains stable. These findings underscore WAVE's robustness in effectively condensing and integrating knowledge from diverse ancestor models.

\subsection{Initialization of Deeper Models}
\label{app:deeper}
We extend our experiments (Table~\ref{tab:lenth}) to deeper models with 24 and 36 layers across different widths ($W_{192}$, $W_{384}$, and $W_{768}$). Table~\ref{tab:deeper} shows that WAVE consistently outperforms He-Init~\cite{chen2021empirical} and TLEG~\cite{xia2024transformer}, achieving higher accuracy across all settings.

For 24-layer models, WAVE surpasses TLEG by 0.8\%, 0.7\%, and 1.8\%, with even greater improvements at 36 layers, confirming its effectiveness in deeper architectures. 
These results demonstrate WAVE’s ability to maintain initialization quality as model depth increases, leveraging size-agnostic weight templates to ensure stable parameter inheritance and robust generalization across variable depths and widths. This scalability establishes WAVE as an efficient initialization strategy for large-scale models.

\section{Additional Analysis}
\subsection{Instincts}
Instincts are natural abilities in organisms, brought by genes, that enable quick adaptation to environments with minimal or even no interaction~\cite{seung2012connectome}. GRL~\cite{feng2023genes} first defines instincts in RL agents, showing that newborn agents can move toward rewards unconsciously. ECO~\cite{feng2024transferring} further extends this definition to supervised learning, demonstrating that networks can quickly classify images with minimal gradient descents, even with a substantial proportion of randomly initialized neurons.

Following the definition of instincts in ECO~\cite{feng2024transferring}, we demonstrate that weight templates, as a new form of the learngene, provide neural networks with strong instincts. As shown in Figure~\ref{fig:app1}, WAVE exhibits stronger initial classification ability compared to other learngenes (including Heur-LG, Auto-LG, TLEG), even after just one epoch of training. This is attributed to the structured size-agnostic knowledge encapsulated in WAVE's weight templates.

\subsection{Strong Learning Ability}
Just as biological instincts enhance learning abilities in organisms, the learning abilities of neural networks are also enhanced by the instincts brought by learngenes.

\begin{figure*}[tb]
  \centering
  \includegraphics[width=\linewidth]{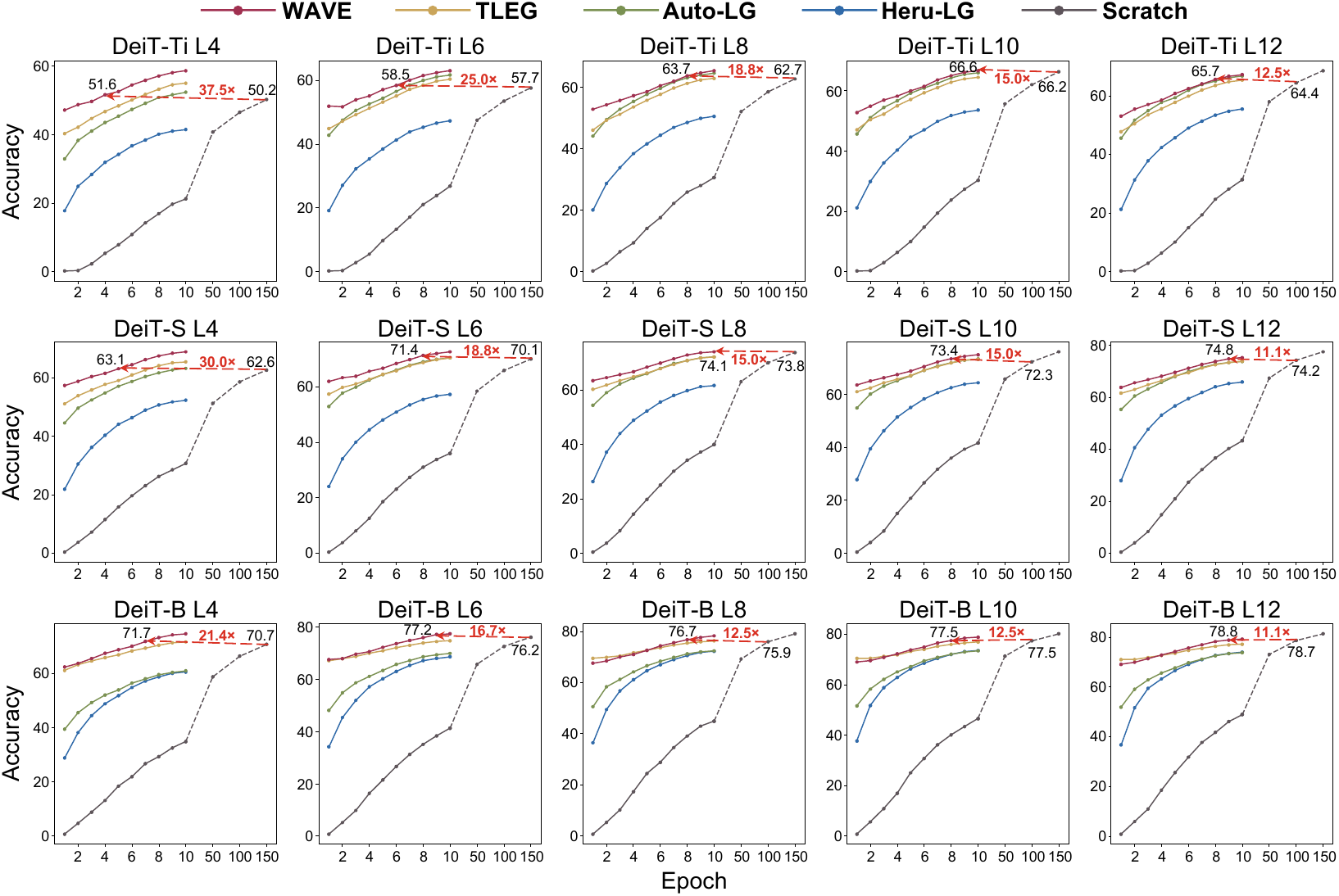}
  \caption{Performance comparisons on ImageNet-1K among WAVE and other learngene methods.}
  \label{fig:app1}
\end{figure*}

Figure~\ref{fig:app1} records the classification accuracy of different learngene methods (10 epochs) and models trained from scratch (150 epochs).
We can see that WAVE outperforms other learngenes (including Heur-LG, Auto-LG and TLEG) and significantly improves training efficiency. 

Compared with the networks trained from scratch, the WAVE-initialized neural networks achieve comparable performance to the neural networks trained from scratch with 150 epochs even after only one epoch of training. 
Taking the DeiT (-Ti, S and B) of 12 layers as an example, the WAVE reduces the training costs around 11$\times$ compared to training from scratch, and such training efficiency is more pronounced in smaller models (37.5$\times$ in DeiT-Ti $L_4$).

\begin{figure*}[tb]
  \centering
  \includegraphics[width=\linewidth]{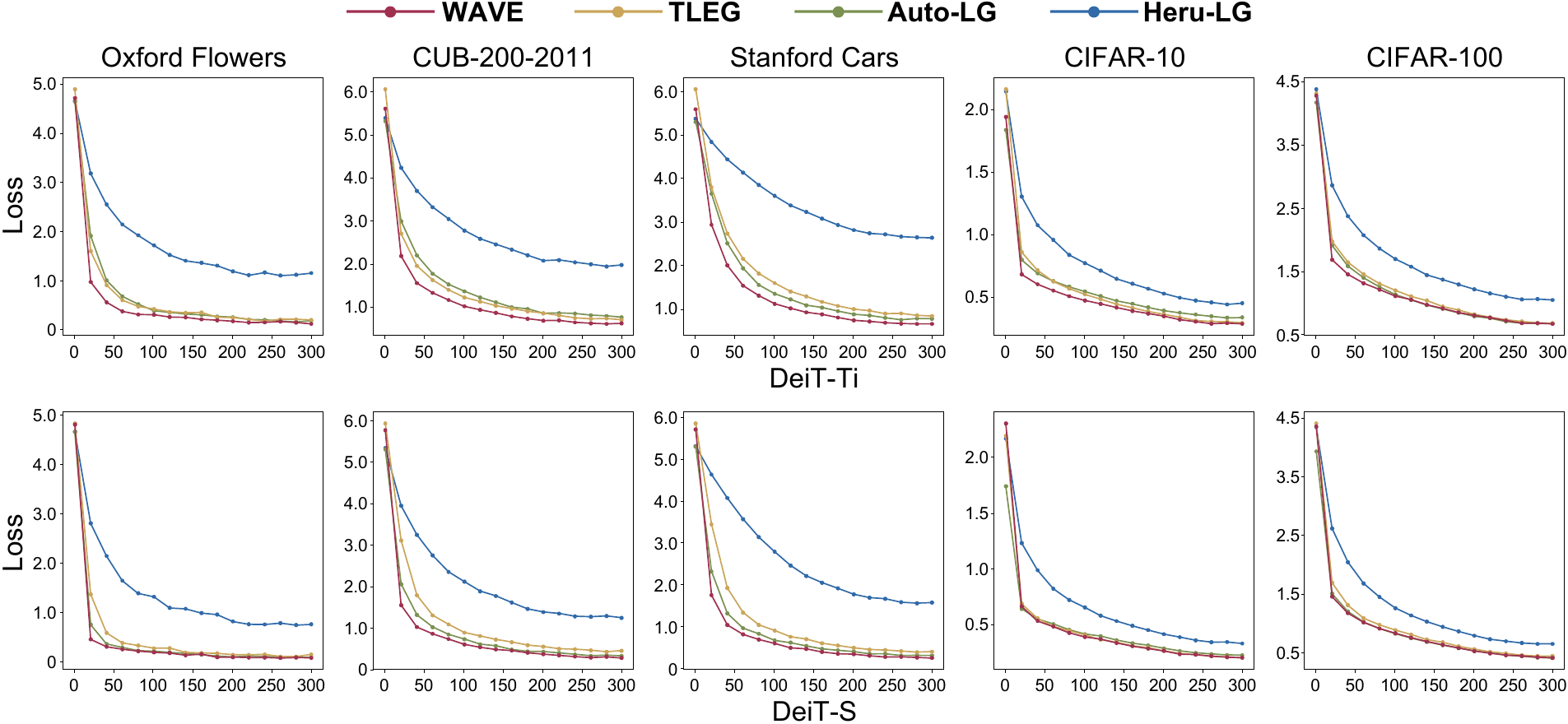}
  \caption{Performance comparisons on small and medium downstream datasets among WAVE and other learngene methods.}
  \label{fig:app2}
\end{figure*}

Such strong learning ability is also evident in models initialized by WAVE on downstream datasets. We visualize the curve of training loss on small and medium datasets (i.e., Oxford Flowers, CUB-200-2011, Stanford Cars, CIFAR-10 and CIFAR-100). As shown in Figure~\ref{fig:app2}, the models initialized by WAVE show faster loss reduction, indicating enhanced learning ability in downstream datasets.

\clearpage

\end{document}